%% file: ms.tex
\documentclass{article}

\PassOptionsToPackage{numbers, sort&compress}{natbib}

\usepackage[preprint]{neurips_2020}

\usepackage[utf8]{inputenc}
\usepackage[T1]{fontenc}
\usepackage{url}
\usepackage{microtype}
\usepackage{graphicx}
\usepackage{booktabs} 
\usepackage{xcolor}
\usepackage{amsfonts}
\usepackage{nicefrac}
\usepackage{algorithm}
\usepackage{algorithmic}
\usepackage{amsmath}
\usepackage{amssymb}
\usepackage{textcomp}
\usepackage{wrapfig}
\usepackage{subcaption}
\usepackage{tikz}
\usetikzlibrary{arrows.meta,calc,decorations.markings,math,arrows.meta}

\usepackage{hyperref}

\newcommand\blfootnote[1]{%
  \begingroup
  \renewcommand\thefootnote{}\footnote{#1}%
  \addtocounter{footnote}{-1}%
  \endgroup
}

\input{notation.tex}

\title{Pruned Neural Networks Are Surprisingly Modular}

\author{
  Daniel Filan$^{\ast1}$ \hskip1.8em Shlomi Hod$^{\ast2}$ \hskip1.8em Cody Wild$^1$ \hskip1.8em Andrew Critch$^1$ \hskip1.8em Stuart Russell$^1$ \\
  $^1$University of California, Berkeley \hskip1em $^2$Boston University \\
  {\small \texttt{\{daniel\_filan, codywild, critch, russell\}@berkeley.edu}} \hskip1em {\small \texttt{shlomi@bu.edu}}
}

\begin{document}

\maketitle
\blfootnote{\!\!$^*$ Equal contributions, order determined randomly.}

\vspace{-3em}

{\Huge \textcolor{red}{This paper has been superceded by arxiv:2103.03386 and arxiv:2110.08058!}}

\begin{abstract}

The learned weights of a neural network are often considered devoid of scrutable internal
structure.
To 
discern structure in these weights, we introduce a measurable notion of modularity for multi-layer perceptrons (MLPs), and investigate the modular structure of MLPs trained on datasets of small images.
Our notion of modularity comes from the graph clustering literature: a ``module'' is a set of neurons with strong internal connectivity but weak external connectivity.
We find that training and weight pruning produces MLPs that are more modular than randomly initialized ones, and often significantly more modular than random MLPs with the same 
(sparse)
distribution of weights.
Interestingly,
they are much more modular when trained with dropout.
We also present exploratory analyses of the importance of different modules for performance and how modules depend on each other.
Understanding the modular structure of neural networks, when such structure exists, will hopefully render their inner workings more interpretable to engineers.\footnote{The source code for this paper is available at \url{https://github.com/prunednnsurprisinglymodular-neurips20/nn_modularity}.} \textcolor{red}{Note that this paper has been superceded by ``Clusterability in Neural Networks'', arxiv:2103.03386 and ``Quantifying Local Specialization in Deep Neural Networks'', arxiv:2110.08058!}
\end{abstract}

\section{Introduction}
\label{intro}

\input{main_sections/intro}

\section{Clustering neural networks}
\label{sec:clustering_NNs}

\input{main_sections/clustering_nns}

\section{Network clusterability results}
\label{sec:clusterability}

\input{main_sections/clust_results}

\section{Module importance and dependencies}
\label{sec:dependency_structure}

\input{main_sections/module_importance_deps}

\section{Related work}
\label{sec:related_work}

\input{main_sections/related_work}

\section{Conclusion}
\label{sec:discussion}

\input{main_sections/conclusion}

\section*{Broader impact}
\label{sec:broader_impact}

\input{main_sections/broader_impact}

\begin{ack}
The authors would like to thank Open Philanthropy for their financial support of this research, the researchers at UC Berkeley's Center for Human-Compatible AI for their advice, and anonymous reviewers for their contributions to improving the paper. Daniel Filan would like to thank Paul Christiano, Rohin Shah, Matthew `Vaniver' Graves, and Buck Shlegeris for valuable discussions that helped shape this research, as well as Andrei Knyazev for his work in helping debug scikit-learn's implementation of spectral clustering and Stephen Casper for his advice during the writing of the paper. Shlomi Hod would like to thank Dmitrii Krasheninnikov for fruitful conversations throughout the summer of 2019.
\end{ack}

\bibliographystyle{plainnat}
\bibliography{nn_modularity}

\input{supplementary.tex}

\end{document}

%% file: notation.tex
\newcommand{\ncut}{\textrm{n-cut}}
\newcommand{\vol}{\textrm{vol}}
\newcommand{\Lnorm}{L_\textrm{norm}}

\newcommand{\R}{\mathbb{R}}

\newcommand{\ts}[1]{\textsuperscript{#1}}

%% file: main_sections/intro.tex
\begin{wrapfigure}{r}{0.45\textwidth}
\vspace{-0.25in}
\begin{center}
\centerline{\includegraphics[width=0.42\textwidth]{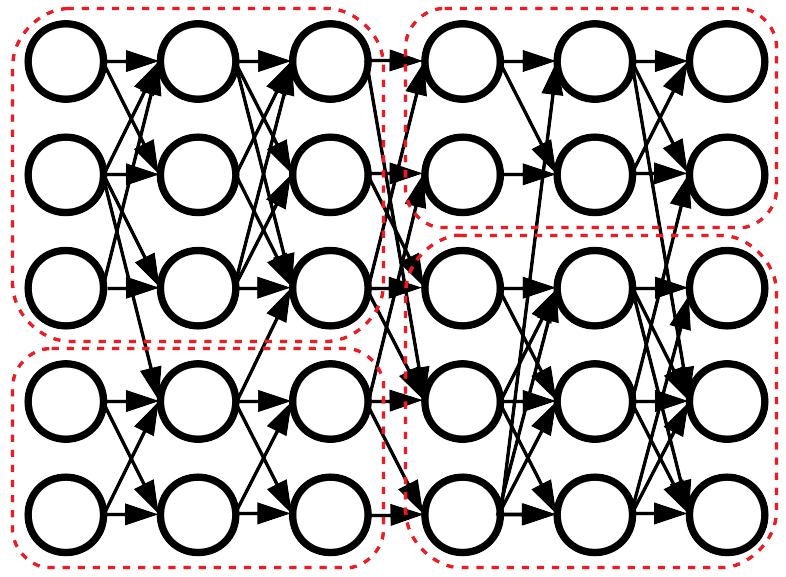}}
\caption{A surprisingly modular pruned neural network, split into modules.}
\label{fig:clustered_net}
\end{center}
\vskip -0.4in
\end{wrapfigure}

Modularity is a common property of biological and engineered systems \cite{clune2013evolutionary, baldwin2000design, booch2007object}.
Reasons given for modularity include adaptability 
and
the ability to handle different situations with common sub-problems. 
It is also desirable from a perspective of transparency:
modular systems allow those analyzing the system to inspect the function of individual modules, and combine their understanding of individual modules into an understanding of the entire system.

Neural networks are composed of distinct layers that are modular in the sense that when constructing a network, different layers are individually configurable.
However, the layers are not modular from the network science perspective which requires modules to be highly internally connected \cite{wagner2007road} (since neurons inside one layer are not directly connected to each other), nor are they modular by the dictionary definition\footnote{\citet{wiktionary-modularity} defines the word `modular' as used in this paper as ``[c]onsisting of separate modules; especially where each module performs or fulfills some specified function and could be replaced by a similar module for the same function, independently of the other modules.''} due to their lack of specific legible functionality.\footnote{That is, beyond the little that can be glimpsed from the kernels of convolutional layers.}

In this work, we develop a measurable notion of the degree of modularity of neural networks: 
roughly, a network is modular to the extent that it can be partitioned into sets of neurons where each set is strongly internally connected, 
but only weakly connected to other sets.
This definition refers only to the \emph{learned weights} of the network, 
not the training data, 
nor the distribution of outputs or activations of the model, 
meaning that it can be analyzed independently of any particular distribution.
More specifically, we use a graph clustering algorithm to decompose trained networks into clusters, and consider each cluster a `module'. We then conduct an empirical investigation into the modularity structure of small MLPs trained on small image datasets.

This investigation shows that networks trained with a final phase of weight-based pruning \cite{han2015learning} are somewhat modular, and are often more modular than approximately 99.7\% of random networks with the same sparsity and distribution of weights.
We also find that networks trained with dropout are much more modular.
We see
some 
modularity when networks increase their accuracy over training,
but much less when they train without increasing their accuracy, as happens 
when 
training on random data.
Finally, we perform a preliminary investigation into the importance and dependency structure of the different modules.

We conclude that the process of learning via gradient descent and pruning selects for modularity.

In section~\ref{sec:clustering_NNs}, we define our notion of modularity and our measure thereof.
We then describe in section~\ref{sec:clusterability} the degree of modularity of networks trained in different fashions,
and discuss experiments done to evaluate the importance of and relationships between different modules in section~\ref{sec:dependency_structure}.
Section~\ref{sec:related_work} gives an introduction to some areas of research related to this paper,
and section~\ref{sec:discussion} contains a summary of our findings and a list of future directions.

%% file: main_sections/clustering_nns.tex
In this section, we provide a graph-theoretic definition of network modularity, and then give an overview of the algorithm we use to measure it, drawing heavily from \citet{von2007tutorial}.

\subsection{Basic definitions}
\label{subsec:clustering_NNs_definitions}

We represent a neural network as a weighted, undirected graph $G$.
To do this, we identify each neuron 
having 
any incoming or outgoing non-zero weights\footnote{When networks are pruned, often some neurons have all incident weights pruned away, leaving them with no functional role in the network. To improve the stability of the clustering algorithm, we ignore such neurons.}, 
including the pixel inputs and logit outputs,
with an integer between $1$ and $N$, where $N$ is the total number of neurons, and take the set of neurons to be the set $V$ of vertices in $G$. Two neurons have an undirected edge between them if they are in adjacent layers, and the weight of the edge will be equal to the absolute value of the weight of the connection between
the two neurons.
We represent the set of weights by the adjacency matrix
$A$ defined by $A_{ij} = A_{ji} :=$ the edge weight between neurons $i$ and $j$. If there is no edge between $i$ and $j$, then $A_{ij} = A_{ji} := 0$. As such, $A$ encodes all the weight matrices of the neural network, but not the biases.

The degree of a neuron $i$ is defined by $d_i := \sum_j A_{ij}$. The degree matrix $D$ is a diagonal matrix where the diagonal elements are the degrees: $D_{ii} := d_i$. 
We define the volume of a set of neurons $X \subseteq V$ as $\vol(X) := \sum_{i \in X} d_i$, 
and the weight between two disjoint sets of neurons $X,Y \subseteq V$ as $W(X,Y) := \sum_{i \in X, j \in Y} A_{ij}$. If $X \subseteq V$ is a set of neurons, then we denote its complement as $\bar{X} := V \setminus X$.

A partition of the network is a collection of subsets $X_1, \dotsc, X_k \subseteq V$ that are disjoint , i.e.\ $X_i \cap X_j = \emptyset$ if $i \neq j$, 
and whose union forms the whole vertex set, $\cup_i X_i = V$. 
Our `goodness measure' of a partition is the normalized cut metric \cite{shi2000normalized} defined as $\ncut(X_1, \dotsc, X_k) := \sum_{i = 1}^k W(X_i, \bar{X_i}) / \vol(X_i)$, which we call `n-cut' in text.\footnote{Note that this differs from the standard definition by a factor of 2.}
This metric will be low if neurons in the same partition element tend to share high-weight edges and those in different partition elements share low-weight edges or no edges at all,
as long as the sums of degrees of neurons in each partition element are roughly balanced. For a probabilistic interpretation of n-cut that gives more intuitive meaning to the quantity, see appendix \ref{app:ncut_meaning}.

Finally, the graph Laplacian 
is defined as $L := D - A$, 
\footnote{For the connection to the second derivative operator on $\mathbb{R}^n$, see \citet{laplacian_explanation} and \citet{von2007tutorial}.} 
and the normalized Laplacian as $\Lnorm := D^{-1}L$. $\Lnorm$ is a positive semi-definite matrix with $N$ real-valued non-negative eigenvalues \cite{von2007tutorial}.

\subsection{Spectral clustering}
\label{subsec:spectral_clustering}

To estimate the `clusterability' of a graph, we use a spectral clustering algorithm to compute a partition, which we will call a clustering, and evaluate the n-cut of that clustering.
Roughly speaking, the spectral clustering algorithm we use \cite{shi2000normalized} solves a relaxation of the problem of finding a clustering that minimizes the n-cut \cite{von2007tutorial}.
It does this by taking the $k$ eigenvectors of $\Lnorm$ with the least eigenvalues, and using them to embed each vertex into $\R^k$---since each eigenvector of $\Lnorm$ is $N$-dimensional, having one real value for every vertex of the graph---then using $k$-means clustering on the
embedded vertices.
It is detailed in algorithm~\ref{alg:spectral_clustering}, which is adapted from \citet{von2007tutorial}. We use the scikit-learn implementation \cite{scikit-learn} using the LOBPCG eigenvalue solver with AMG preconditioning \cite{knyazev2001toward, borzi2006algebraic}.

\begin{algorithm}[tb]
\caption{Normalized Spectral Clustering \cite{shi2000normalized}}
\label{alg:spectral_clustering}
\begin{algorithmic}
\STATE {\bfseries Input:} Adjacency matrix $A$, number $k$ of clusters
\STATE Compute the normalized Laplacian $\Lnorm$
\STATE Compute the first $k$ eigenvectors $u_1, \dotsc, u_k \in \R^N$ of $\Lnorm$
\STATE Form the matrix $U \in \R^{k \times N}$ whose $j$\ts{th} row is $u_j^\top$
\STATE For $n \in \{1, \dotsc, N\}$, let $y_n \in \R^k$ be the $n$\ts{th} column of $U$
\STATE Cluster the points $(y_n)_{n=1}^N$ with the $k$-means algorithm into clusters $C_1, \dotsc, C_k$
\STATE {\bfseries Return:} Clusters $X_1, \dotsc, X_k$ with $X_i = \{n \in \{1, \dotsc, N\} \mid y_n \in C_i \}$
\end{algorithmic}
\end{algorithm}

We will define the n-cut of a network as the n-cut of the clustering that algorithm~\ref{alg:spectral_clustering} returns. As such, since the n-cut is low when the network is clusterable or modular, we will describe a decrease in n-cut as an increase in modularity or clusterability, and vice versa.\footnote{We will use the words `clusterable' and `modular' more or less interchangeably. Despite our informal use of `modularity' here, note that modularity also has a technical definition in the network science literature \cite{newman2006modularity}, which we are not using.}

%% file: main_sections/clust_results.tex
In this section, we report the results of experiments designed to determine the degree of clusterability of trained neural networks. 
Unless otherwise specified, for each experiment we train an MLP with 4 hidden layers, each of width 256, for 20 epochs of Adam \cite{kingma2014adam} with batch size 128. 
We then run weight pruning on a polynomial decay schedule \cite{zhu2017prune} up to 90\% sparsity for an additional 20 epochs. Pruning is used since 
the pressure to minimize connections plausibly causes modularity in biological systems \cite{clune2013evolutionary}.
For further details, 
see appendix~\ref{app:train_hypers}.

Once the network is trained, we convert it into a graph. 
We then run algorithm~\ref{alg:spectral_clustering} with 4 clusters (i.e.\ $k=4$),\footnote{We choose 4 to avoid the partitioning being overly simple while minimizing the computational budget, and because we expect higher-degree clusterability to be reflected in 4-way clusterability. 
Appendix~\ref{app:n_clusters} shows similar results for 7 and 10 clusters, but different results for 2 clusters.
} and evaluate the n-cut of the resulting clustering. 
Next, we sample 320 random networks by randomly shuffling the weight matrix of each layer of the trained network.
We convert these networks to graphs, cluster them, and find their n-cuts. 
We then compare the n-cut of the trained network to the sampled n-cuts, estimating the one-sided $p$-value \cite{north2002note}.
This determines whether the trained network is more clusterable than one would predict based only on its sparsity and distribution of weights.

We use three main datasets: MNIST \cite{mnist}, Fashion-MNIST \cite{fashion_mnist}, and CIFAR-10 \cite{cifar}. MNIST and Fashion-MNIST are used as-is, while CIFAR-10 is down-sampled to 28$\times$28 pixels and grayscaled to fit in the same format as the other two. MNIST and Fashion-MNIST each have 60,000 training examples and 10,000 test examples, while CIFAR-10 has 50,000 training examples and 10,000 test examples. On CIFAR-10, we prune for 40 epochs rather than 20 for improved accuracy.

\begin{figure}[tb]
 \begin{center}
             \centerline{\includegraphics[width=0.6\textwidth]{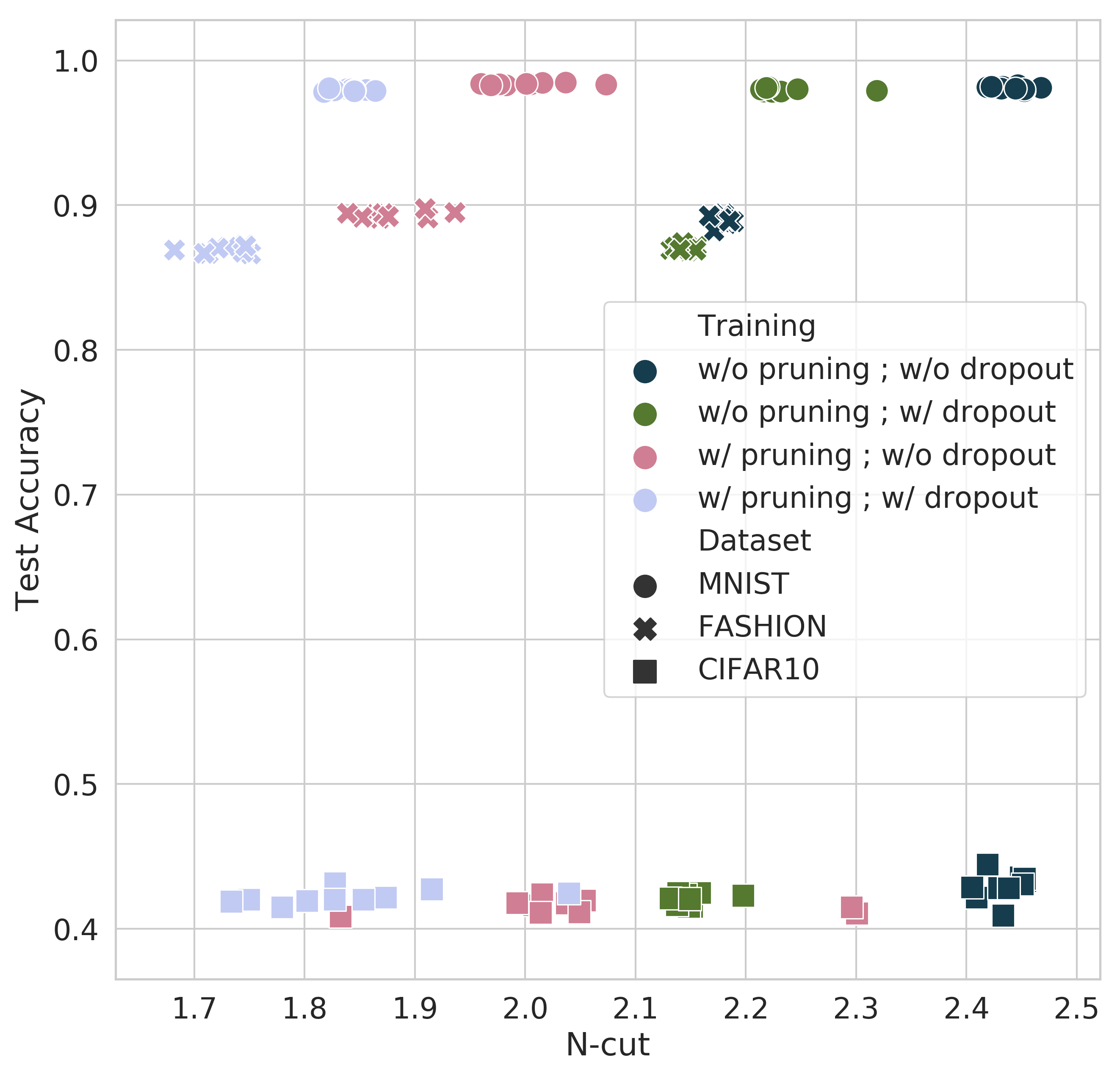}}
    \caption{Scatter plot of each trained network's n-cut and accuracy, evaluated both just before pruning (in dark blue and green) and at the end of training (in light blue and pink). For a given dataset and training method, n-cut appears to be roughly independent of accuracy. 
Note 
the reliable drop in n-cut from pruning and dropout, and the large spread of values for the n-cuts of models trained on CIFAR-10.
}
\label{fig:ncuts_accs}
 \end{center}
 
\end{figure}

\begin{figure}[th]
 \begin{center}
 \centerline{\includegraphics[width=0.8\textwidth]{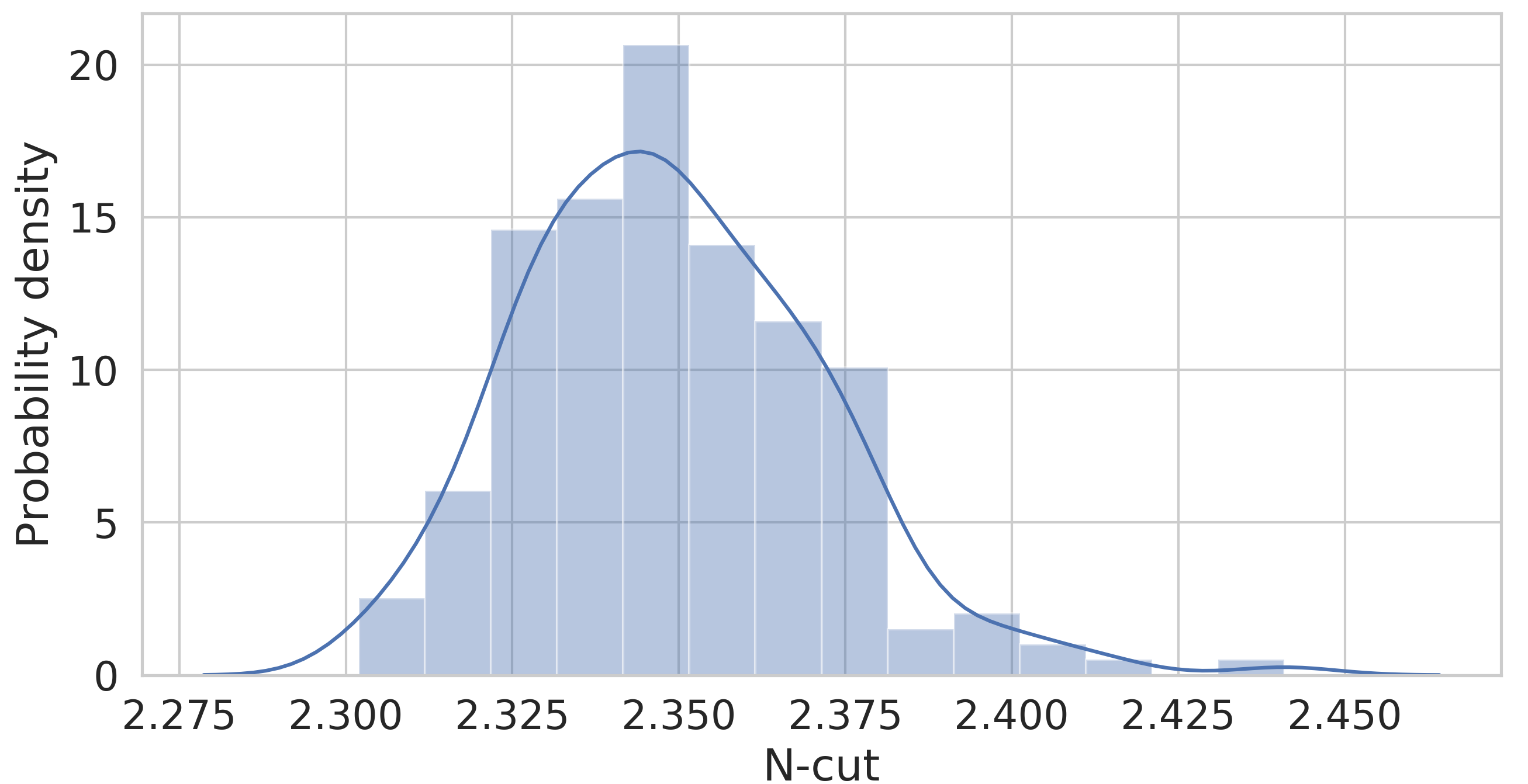}}
 \caption{
 N-cuts of 200 randomly-initialized networks---histogram and kernel density estimate.
 Produced by the \texttt{distplot} function of seaborn 0.9.0 \cite{seaborn_0_9_0} with default arguments.}
 \label{fig:init_ncuts}
\end{center}
\end{figure}

We train 10 networks on each dataset. Our results are shown in table~\ref{tab:ncut-stats-basic}. As can be seen, we train to approximate test accuracies of 98\% on MNIST, 89\% on Fashion-MNIST, and 42\% on CIFAR-10.
The pruned networks are often more clusterable than all 320 random shufflings.
The frequency of this depends on the dataset: every network trained on Fashion-MNIST had a lower n-cut than all 320 random permutations, while only two-thirds of networks trained on MNIST did, and only 1 out of 10 networks trained on CIFAR-10 did. In fact, 5 of the 10 networks trained on CIFAR-10 were \emph{less} clusterable than all their own shuffles. More complete data is available in appendix~\ref{app:clusterability_data}, and results for the n-cuts and test accuracies of all trained networks both pre-{} and post-pruning (as well as when trained with dropout, see subsection~\ref{subsec:dropout_clust}) are shown in figure~\ref{fig:ncuts_accs}.
To aid in the interpretation of these n-cut values, the n-cuts of 200 randomly-initialized networks are shown in figure~\ref{fig:init_ncuts}. 
The support lies entirely between 2.3 and 2.45.\footnote{
Note that n-cut is invariant to an overall scaling of the weights of a graph, and therefore changes from initialization are not simply due to the norm of the weights vector increasing or decreasing.
}

\begin{table*}[tb]
\caption{
Clusterability results. 10 networks were trained on each dataset without dropout, and 10 were trained with dropout.
``Fashion'' is short for Fashion-MNIST. ``Prop.\ sig.'' means the proportion of trained networks out of 10 that were significantly clusterable at the $p<1/320$ level (ignoring Bonferroni corrections)---i.e.\ had a lower n-cut than all random shuffles of their weight matrices. Training and test accuracies reported are the means over the 10 trained networks. `N-cuts' gives the sample mean and standard deviation of the 10 trained networks. `Dist.\ n-cuts' gives the sample mean and standard deviation over all shuffles of all 10 trained networks.
}
\label{tab:ncut-stats-basic}
\vskip 0.15in
\begin{center}
\begin{small}
\begin{tabular}{lccccccc}
\toprule
Dataset & Dropout & Prop.\ sig.\ & Train acc.\ & Test acc.\ &  N-cuts & Dist.\ n-cuts \\
\midrule
MNIST    & $\times$ & $0.7$ & $1.00 $ & $0.984$ & $2.000 \pm 0.035$ & $2.042 \pm 0.017$ \\
MNIST    & $\surd$ & $1.0$ & $0.967$ & $0.979$ & $1.840 \pm 0.015$ & $2.039 \pm 0.019$ \\
Fashion  & $\times$ & $1.0$ & $0.983$ & $0.893$ & $1.880 \pm 0.030$ & $1.992 \pm 0.018$ \\
Fashion  & $\surd$ & $1.0$ & $0.863$ & $0.869$ & $1.726 \pm 0.022$ & $2.013 \pm 0.017$ \\
CIFAR-10 & $\times$ & $0.1$ & $0.650$ & $0.415$ & $2.06  \pm 0.14 $ & $2.001 \pm 0.017$ \\
CIFAR-10 & $\surd$ & $0.9$ & $0.427$ & $0.422$ & $1.840 \pm 0.089$ & $1.997 \pm 0.014$ \\
\bottomrule
\end{tabular}
\end{small}
\end{center}
\vskip -0.1in
\end{table*}

It should be noted that the trained networks are still rather monolithic in an absolute sense---just less so than initialized networks or networks with the same distribution of weights, hinting that the learning process may be promoting modularity.

\subsection{How does dropout affect clusterability?}
\label{subsec:dropout_clust}

Dropout is a method of training networks that reduces overfitting 
by randomly setting neuronal activations to zero during training \cite{srivastava2014dropout}. One might expect this to decrease clusterability, since it encourages neurons to place weight on a large number of previous neurons, rather than relying on a few that might get dropped out. However, we see the reverse.

We run exactly the same experiments as described earlier, but applying dropout to all layers other than the output with a rate of 0.5 for MNIST and Fashion-MNIST, and 0.2 for CIFAR-10.\footnote{The dropout rate of 0.2 was chosen for CIFAR-10 because the network failed to learn when the rate was 0.5.}
As shown in table~\ref{tab:ncut-stats-basic}, we consistently find statistically significantly low n-cuts, 
and the means and standard deviations of the n-cuts of dropout-trained networks are lower than those of the networks trained without dropout.\footnote{The Z-statistics of the n-cut distributions with and without dropout training are $\sim13$ for MNIST and Fashion-MNIST and $\sim 4$ for CIFAR-10. Since the underlying distributions may well not be normal, these numbers should be treated with caution.}
As figure~\ref{fig:ncuts_accs} shows, there is essentially no overlap between the distributions of n-cuts of networks trained with and without dropout, holding fixed the training dataset and whether or not pruning has occurred.

\subsection{Does clusterability come from training alone?}
\label{subsec:random_dataset}

There are two potential dataset-agnostic explanations for why clusterability would increase during training. 
The first is that it increases naturally as a byproduct of applying stochastic gradient descent (SGD) updates. 
The second is that in order to accurately classify inputs, networks adopt relatively clusterable structure. 
To distinguish between these two explanations, we run three experiments on 
a dataset of 28$\times$28 images with iid uniformly random pixel values, associated with random labels between 0 and 9.

In the \textbf{unlearnable random dataset experiment}, we train on 60,000 random images with default hyperparameters, 10 runs with dropout and 10 runs without. 
Since the networks are unable to memorize the labels, this tests the effects of SGD controlling for accuracy.
We compare the n-cuts of the unpruned networks against the distribution of randomly initialized networks to check whether SGD without pruning increases clusterability.
We also compare the n-cuts from the pruned networks against the distribution of n-cuts from shuffles of those networks, to check if SGD increases clusterability in the presence of pruning more than would be predicted purely based on the increase in sparsity.

In the \textbf{kilo-epoch random dataset experiment}, we modify the unlearnable random dataset experiment to remove the pruning and train for 1000 epochs instead of 20, 
to check if clusterability simply takes longer to emerge from training when the dataset is random. Note that even in this case, the network is unable to classify the training set better than random.

In the \textbf{memorization experiment}, we modify the random dataset and training method to be more easily learnable. To do this, we reduce the number of training examples to 3,000, train without pruning for 100 epochs and then with pruning for 100 more epochs, and refrain from shuffling the dataset between epochs. As a result, the network is able to memorize the labels only when dropout is not applied, letting us observe whether SGD, pruning, and learning can increase clusterability on an arbitrary dataset without dropout.

As is shown in table~\ref{tab:ncut-stats-random}, the unlearnable random dataset experiment shows no increase in clusterability before pruning relative to the initial distribution shown in figure~\ref{fig:init_ncuts}, suggesting that it is not a result of the optimizer alone. 
We see an increase in clusterability after pruning, but the eventual clusterability is no more than 
that of 
a random network with the same distribution of weights.

\begin{table*}[tp]
\caption{Results from the unlearnable random dataset experiment. Reporting as in table~\ref{tab:ncut-stats-basic}. ``Unp'' is short for unpruned. Accuracies and distributions are of pruned networks.
}
\label{tab:ncut-stats-random}
\vskip 0.15in
\begin{center}
\begin{small}
\begin{tabular}{cccccccc}
\toprule
Dropout & Prop.\ sig.\ & Train acc.\ & Unp.\ n-cuts &  N-cuts & Dist.\ n-cuts \\
\midrule
$\times$ & $0.0$ & $0.102$ & $2.338 \pm 0.030$ & $2.093 \pm 0.023$ & $2.081 \pm 0.014$ \\
$\surd$  & $0.0$ & $0.102$ & $2.323 \pm 0.020$ & $2.053 \pm 0.015$ & $2.061 \pm 0.015$ \\
\bottomrule
\end{tabular}
\end{small}
\end{center}
\vskip -0.1in
\end{table*}

The results from the kilo-epoch random dataset experiment are shown in table~\ref{tab:ncut-stats-random-x50}. 
The means and standard deviations suggest that even a long period of training caused no increase in clusterability relative to the distribution shown in figure~\ref{fig:init_ncuts}, 
while pruning only caused the increase in clusterability via sparsification. 
However, some runs had significantly low n-cut after pruning at 
$p<1/320$, 
suggesting that some outliers were 
indeed 
abnormally clusterable. 
As such, the results of this experiment should be treated as somewhat ambiguous.

\begin{table*}[tp]
\caption{Results from the kilo-epoch random dataset experiment. Reporting as in table~\ref{tab:ncut-stats-random}.
}
\label{tab:ncut-stats-random-x50}
\vskip 0.15in
\begin{center}
\begin{small}
\begin{tabular}{cccccccc}
\toprule
Dropout  & Prop.\ sig.\ & Train acc.\ & Unp.\ n-cuts      &  N-cuts           & Dist.\ n-cuts \\
\midrule
$\times$ & $0.1$        & $0.102$     & $2.342 \pm 0.022$ & $2.082 \pm 0.017$ & $2.082 \pm 0.014$ \\
$\surd$  & $0.3$        & $0.102$     & $2.329 \pm 0.020$ & $2.043 \pm 0.028$ & $2.061 \pm 0.017$ \\
\bottomrule
\end{tabular}
\end{small}
\end{center}
\vskip -0.1in
\end{table*}

The results of the memorization experiment, shown in table~\ref{tab:ncut-stats-random-memorization}, are different for the networks trained with and without dropout. 
Networks trained with dropout 
did not memorize the dataset, and 
seem to have n-cuts in line with the random distribution, although 3 of them had statistically significantly low n-cuts. 
Those trained without dropout did memorize the dataset and were all statistically significantly clusterable. 
In fact, their degree of clusterability is similar to that of those trained on the Fashion-MNIST dataset without dropout.
Before the onset of pruning, their n-cuts are not particularly lower than the distribution of those of randomly initialized MLPs shown in figure~\ref{fig:init_ncuts}.

\begin{table*}[t]
\caption{Results from the memorization experiment. Reporting as in table~\ref{tab:ncut-stats-random}.
}
\label{tab:ncut-stats-random-memorization}
\vskip 0.15in
\begin{center}
\begin{small}
\begin{tabular}{cccccccc}
\toprule
Dropout  & Prop.\ sig.\ & Train acc.\ & Unp.\ n-cuts      &  N-cuts           & Dist.\ n-cuts \\
\midrule
$\times$ & $1.0$        & $1.00 $     & $2.464 \pm 0.014$ & $1.880 \pm 0.017$ & $2.008 \pm 0.014$ \\
$\surd$  & $0.3$        & $0.113$     & $2.333 \pm 0.018$ & $2.038 \pm 0.033$ & $2.055 \pm 0.023$ \\
\bottomrule
\end{tabular}
\end{small}
\end{center}
\vskip -0.1in
\end{table*}

Overall, these results suggest that 
the training process promotes modularity as a by-product of learning or memorization, and not automatically.

\subsection{Does clusterability come from topology alone?}
\label{subsec:connectivity_structure}

Since SGD alone does not appear to increase clusterability, one might suppose that 
the increase in clusterability relative to random networks is due to the pruning producing a clusterable topology, and that the values of the non-zero weights are unimportant. To test this, we compare each trained network to a new distribution: instead of randomly shuffling all elements of each weight matrix, we only shuffle the non-zero elements, thereby preserving the network's topology,

\begin{figure}[th]
\begin{center}
\centerline{\includegraphics[width=0.8\textwidth]{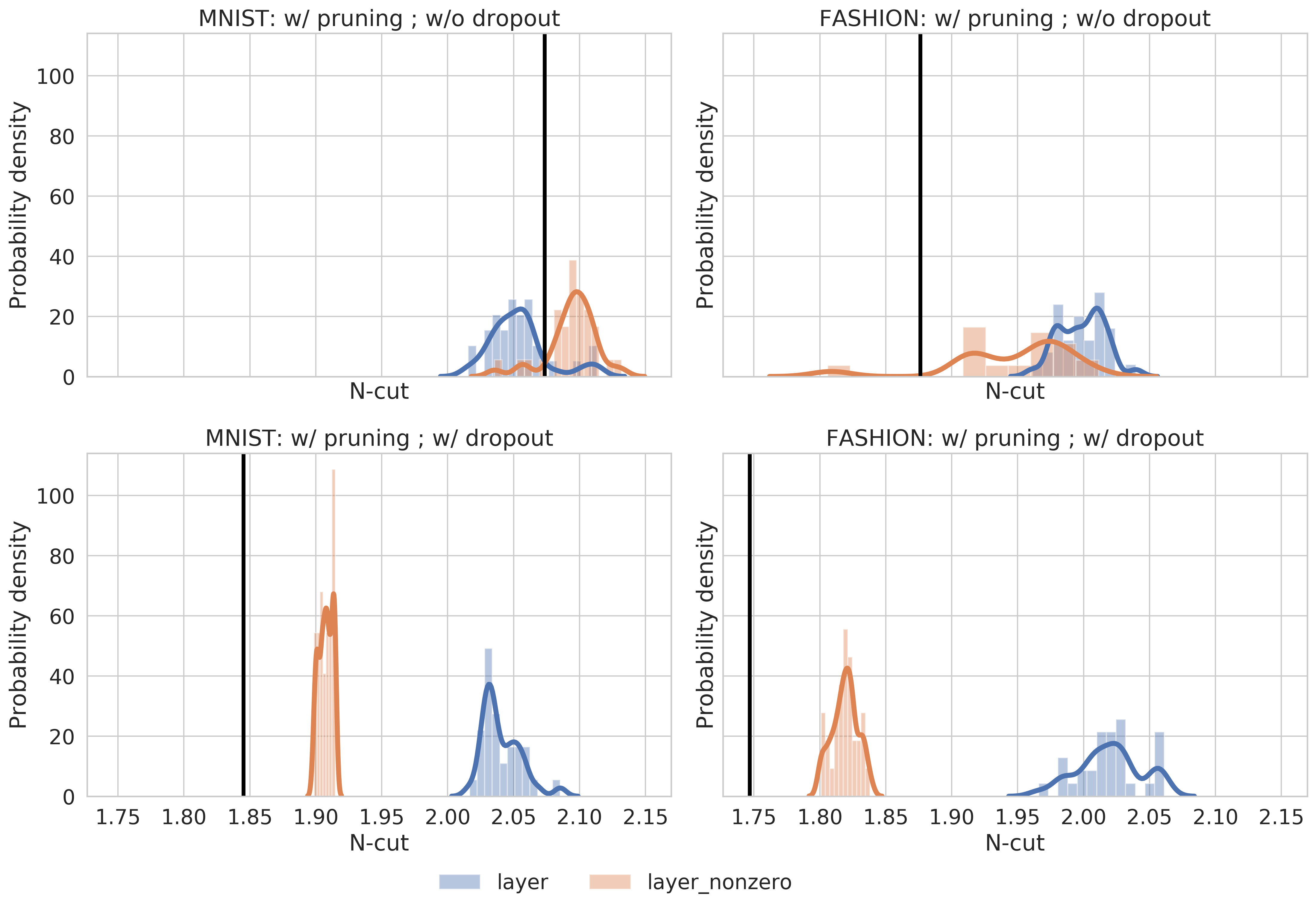}}
\caption{N-cuts of pruned networks trained on MNIST and Fashion-MNIST with and without dropout, compared to the distribution of n-cuts of networks generated by shuffling all elements of each weight matrix (shown in blue, labeled `layer'), as well as the distribution of n-cuts of networks generated by shuffling only the non-zero elements of each weight matrix so as to preserve network topology (shown in orange, labeled `layer-nonzero'). Realized n-cuts are shown as black vertical lines. 
Produced by the \texttt{distplot} function of seaborn 0.9.0 \cite{seaborn_0_9_0} with default arguments.}
\label{fig:ncut_dists}
\end{center}
\end{figure}

Figure~\ref{fig:ncut_dists} shows the n-cuts of some representative networks compared to the distribution of n-cuts of all shuffled networks, and also the distribution of n-cuts of the topology-preserving shuffles. We see three things:
first, that in all cases our networks are more clusterable than would be expected given their topology; 
second, that the topology-preserving shuffles are more clusterable than other shuffles,
suggesting that the pruning process is removing the right weights to promote clusterability;
and third, that with dropout, the distribution of topology-preserving shuffles has much lower n-cuts than the distribution of all shuffles.
More complete data showing the same qualitative results is displayed in figures \ref{fig:violin-pruned-no-dropout} and \ref{fig:violin-pruned-dropout}.

%% file: main_sections/module_importance_deps.tex


We have seen that trained networks tend to be more modular than chance would predict. 
However, it remains to be shown that this topological property has any functional significance.
In this section, we validate the meaningfulness of the clustering by seeing what happens if we lesion clusters---setting the activations of their neurons to 0.

\subsection{Importance: single lesion experiments}
\label{subsec:single_lesion}

\begin{figure}[th]
\begin{center}
\centerline{\includegraphics[width=0.8\textwidth]{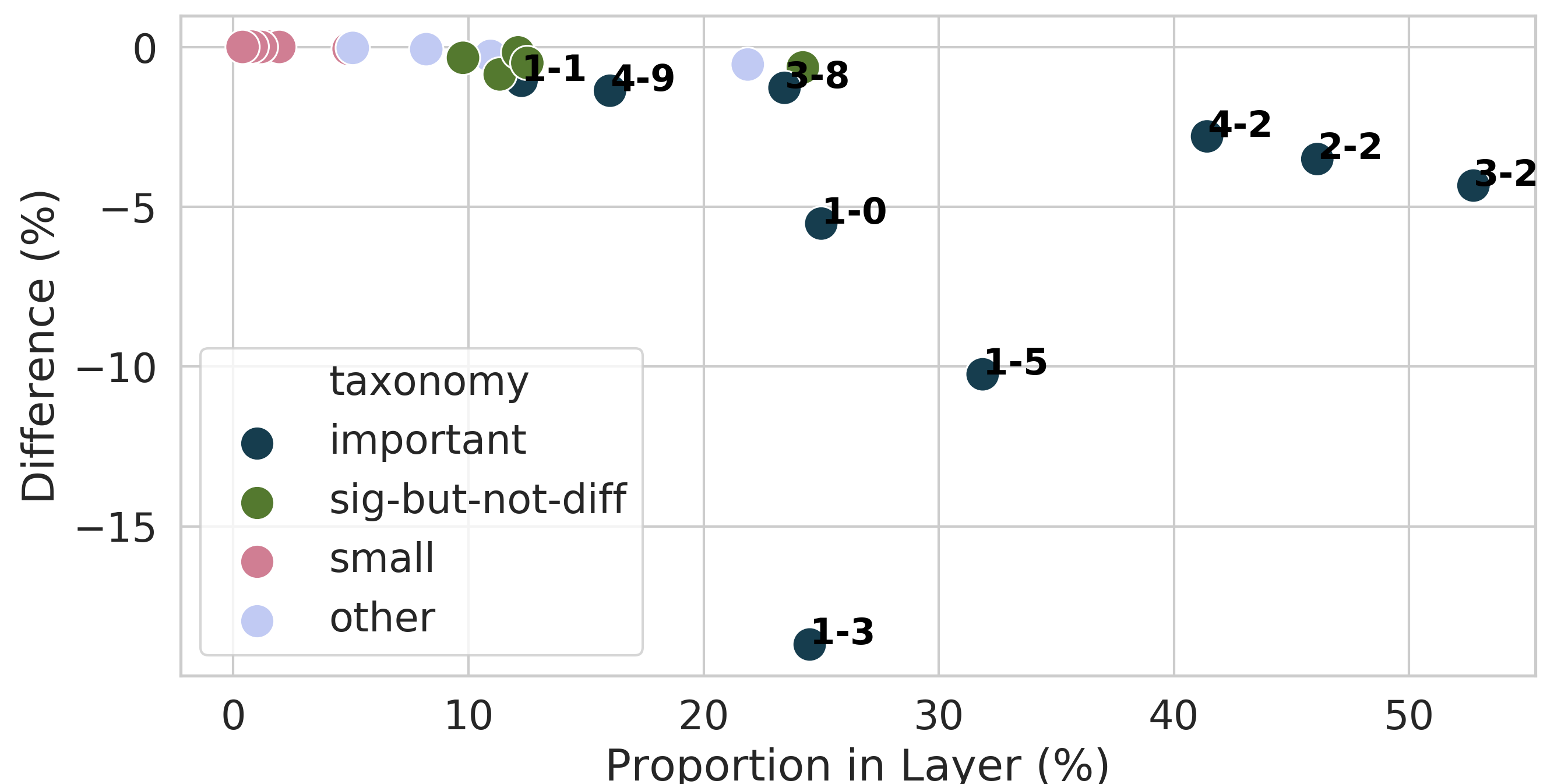}}
\caption{Plot of different sub-modules of a network trained on Fashion-MNIST using pruning and dropout. Important sub-modules are labeled first by their layer number and then their module number. The horizontal axis shows the proportion of their layer's neurons that are in the sub-module, and the vertical axis shows the reduction in accuracy caused by lesioning the sub-module. `important' means that the drop in accuracy is greater than one percentage point, statistically significant compared to the lesioning of a random set of neurons, and that the sub-module consists of 5\% or more of the layer. `sig-but-not-diff' means that the drop in accuracy is significant but less than 1 percentage point, `small' means that the sub-module consists of less than 5\% of the layer, and `other' means that the drop in accuracy is not statistically significant and less than 1 percentage point. Table included in appendix~\ref{app:lesion_experiment_data}.}
\label{fig:single_lesion_plot}
\end{center}
\end{figure}
 
In this subsection 
and the next,
the atomic units of lesioning are the intersections of modules 
with hidden layers, which we will call ``sub-modules''.
This is for two reasons: firstly, because whole modules are large enough that it would be difficult to analyze them, since lesioning any such large set would cause significant damage; and secondly, to control for the difference in the types of representations learned between earlier and later layers.

Here, we investigate the importance of single sub-modules. To do so, for each sub-module, we set all weights incoming to the constituent neurons to 0, while leaving the rest of the network untouched.
We then determine the damaged network's test-set accuracy, and in particular how much lower it is than the accuracy of the whole network.
To assess how meaningful the sub-module is, we use three criteria: firstly, the drop in accuracy should be 
greater than 1 percentage point;
secondly, the drop should not simply be due to the number of damaged neurons;
and thirdly, the sub-module should be at least 5\% of the neurons of the layer.
To evaluate the second criterion, 100 times we randomly pick the same number of neurons as were in the sub-module from the same layer to lesion, and collect the distribution of accuracy drops.
We say that the criterion is met if the actual accuracy drop is greater than all 100 sampled drops.

In figure~\ref{fig:single_lesion_plot}, we show data on the importance of sub-modules of an MLP trained on Fashion-MNIST with dropout that has been clustered into 10 modules.\footnote{The number 10 was chosen to increase the granularity of analysis.} 
Many sub-modules are too small to be counted as important, and many are statistically significantly impactful but not practically significant.
However, some sub-modules clearly are practically important for the functioning of the network.

\subsection{Dependencies: double lesion experiments}
\label{subsec:double_lesion}

\begin{figure}[tb]
    \begin{center}
    \begin{subfigure}{0.48\textwidth}
        \centering
        \includegraphics[width=\textwidth]{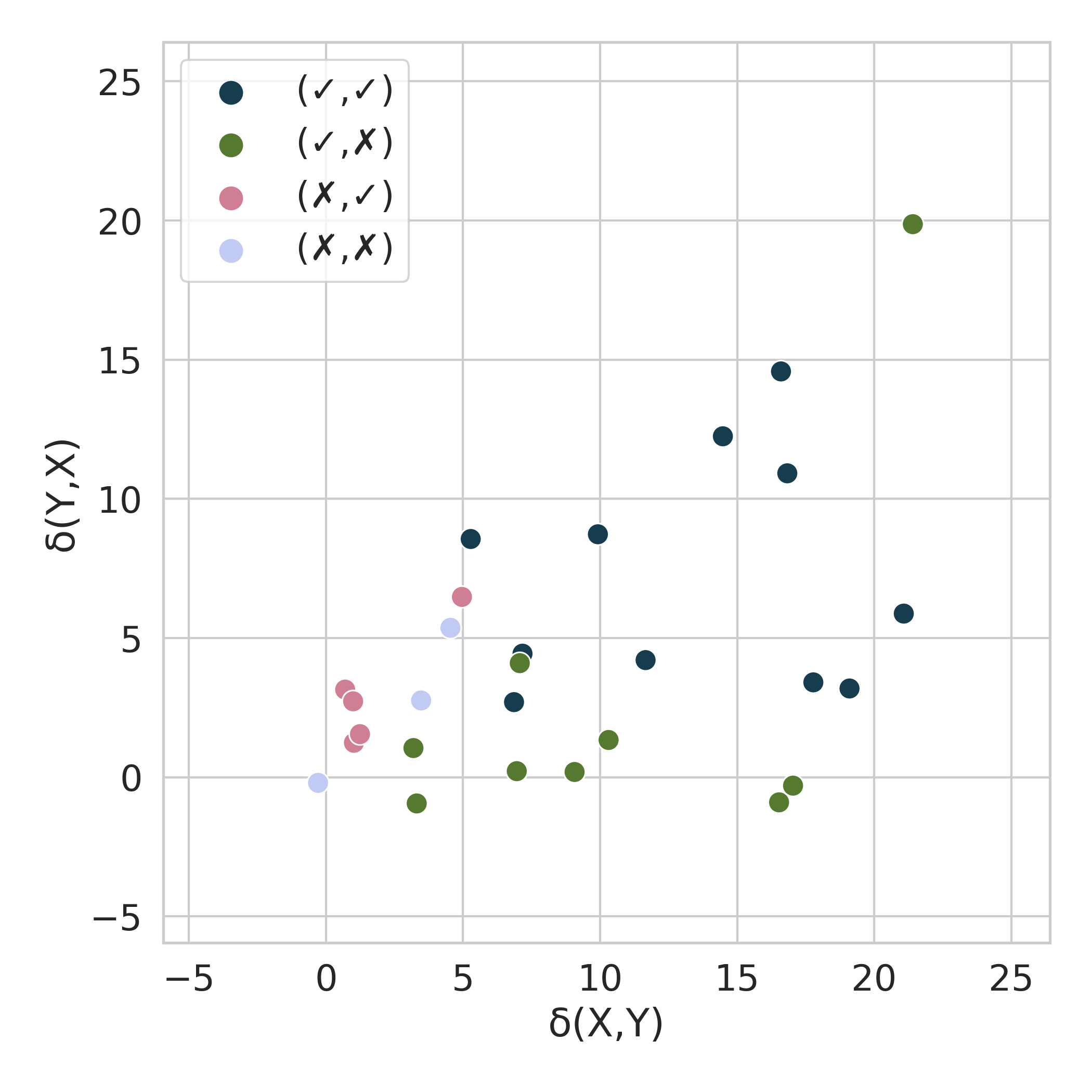} \\
        (a)
    \end{subfigure}
    \hspace*{10pt}
        \begin{subfigure}{0.48\textwidth}
        \begin{center}
            \begin{tikzpicture}
                \begin{scope}[every node/.style={circle,thick,draw}]
                    \node[shape=circle,draw=black] (10) at (0,0) {1-0};
                    \node[shape=circle,draw=black] (11) at (2,0) {1-1};
                    \node[shape=circle,draw=black] (13) at (4,0) {1-3};
                    \node[shape=circle,draw=black] (15) at (6,0) {1-5};
                    \node[shape=circle,draw=black] (22) at (5.4,-1.5) {2-2};
                    \node[shape=circle,draw=black] (32) at (4,-3) {3-2};
                    \node[shape=circle,draw=black] (38) at (0,-3) {3-8} ;
                    \node[shape=circle,draw=black] (42) at (0,-4.5) {4-2} ;
                    \node[shape=circle,draw=black] (49) at (6,-4.5) {4-9} ;
                \end{scope}
                \begin{scope}[>={Stealth[black]},
                              every edge/.style={draw=,very thick}]
                    \path [->](10) edge node[left] {} (38);
                    \path [->](11) edge node[left] {} (22);
                    \path [->](11) edge node[left] {} (38);
                    \path [->](11) edge node[left] {} (42);
                    \path [->](11) edge node[left] {} (49);
                    \path [->](13) edge node[left] {} (38);
                    \path [->](13) edge node[left] {} (49);
                    \path [->](15) edge node[left] {} (22);
                    \path [->](15) edge node[left] {} (38);
                    \path [->](15) edge node[left] {} (49);
                    \path [->](22) edge node[left] {} (32);
                    \path [->](22) edge node[left] {} (42);
                    \path [->](22) edge node[left] {} (49);
                    \path [->](15) edge node[left] {} (49);
                    \path [->](32) edge node[left] {} (42);
                    \path [->](32) edge node[left] {} (49);
                    \path [->](38) edge node[left] {} (42);
                    \path [->](38) edge node[left] {} (49);
                \end{scope}
            \end{tikzpicture} \\
        (b)
        \end{center}
    \end{subfigure}
    \caption{(a) The conditional importances of pairs $(X,Y)$ of important sub-modules and their $\delta$ values in percentage points. $(\surd, \times)$ means that $X$ is important conditioned on $Y$ but not the reverse, and similarly for the other table entries. Data used to generate this plot is in appendix~\ref{app:lesion_experiment_data}. (b) The sub-module dependency graph that we derive from data shown in (a). Sub-modules are identified first by their layer number, then by their module number.}
    \label{fig:dependency_info}
    \end{center}
    \vspace{-0.3in}
\end{figure}
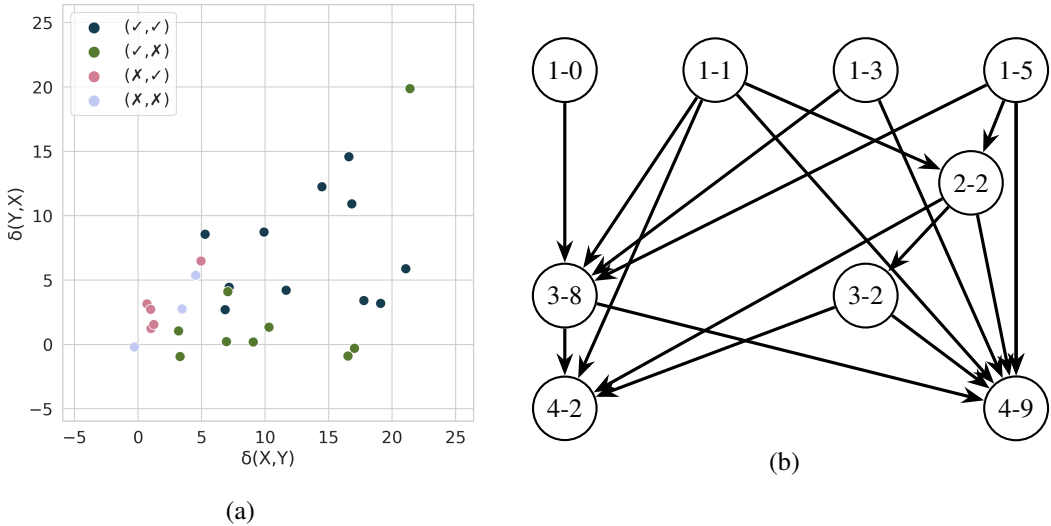

Now that we know which sub-modules are important, 
we attempt
to understand how the important sub-modules depend on each other. To do this, we conduct experiments where we lesion two different important sub-modules, which we'll call $X$ and $Y$, in different layers.
First, we measure the loss in accuracy when both are lesioned, which we'll call $\ell(X \cup Y)$.
We then take 50 random subsets $Y'$ of $|Y|$ neurons from the same layer as $Y$, lesion $X$ and $Y'$, call the loss in accuracy $\ell(X \cup Y')$, and check if $\ell(X \cup Y)$ is larger than all samples of $\ell(X \cup Y')$.
This tests if the damage from lesioning $Y$ is statistically significant given how many neurons are contained in $Y$, and given that we are already lesioning $X$.
We also calculate $\delta(Y,X) := \ell(X \cup Y) - \ell(X)$, which is the additional damage from lesioning $Y$ given that $X$ has been lesioned.
If $\ell(X \cup Y)$ is statistically significantly large, 
and if $\delta(Y,X)$ is larger than 1 percentage point, we say that sub-module $Y$ is important conditioned on sub-module $X$.
Similarly, we test if $X$ is important conditioned on $Y$ 
by comparing $\ell(X \cup Y)$ to the distribution of $\ell(X' \cup Y)$, and by determining the size of $\delta(X,Y)$.
Figure~\ref{fig:dependency_info}~(a) plots the $\delta$ values and importances of different pairs of clusters.
Data for significance of all sub-modules, not merely the important ones, are presented in appendix~\ref{app:lesion_experiment_data}.

By examining the importances of sub-modules conditioned on each other, we can attempt to construct a dependency graph of sub-modules by determining which sub-modules send information to which others.
Consider a pair of sub-modules $(X,Y)$ where $X$ is in an earlier layer than $Y$, and where both are individually important.
If $X$ is no longer important when conditioned on $Y$, we conjecture that all information from $X$ is sent to $Y$, as otherwise lesioning $X$ would cause areas other than $Y$ to mis-fire, reducing accuracy. Similarly, if $Y$ is no longer important conditioned on $X$, we conjecture that all information that is sent to $Y$ comes from $X$, as otherwise other sources of classification-relevant signal would be lost, reducing accuracy.

We therefore
conclude that for any two important sub-modules $X$ and $Y$, if one is unimportant conditioned on the other, then information flows from one to another. This, combined with data shown in figure~\ref{fig:dependency_info}~(a), lets us draw edges in a dependency graph of sub-modules, which is shown in figure~\ref{fig:dependency_info}~(b).
Note that sub-modules of module 2 seem to send information to each other, which is exactly what we would expect if modules were internally connected.

%% file: main_sections/related_work.tex
Previous research has explored modularity in neural networks. \citet{watanabe2018modular} demonstrate a different way of grouping neurons into modules, the interpretation of which is discussed by \citet{watanabe2018understanding} and \citet{watanabe2019interpreting}. \citet{davis2019nif} cluster small networks based on statistical properties, rather than on the weights of the network, 
and \citet{lu2019checking} bi-clusters neurons in a hidden layer and training data using neural attribution methods.
More broadly, \citet{clune2013evolutionary} discuss modularity in biological networks, and give an overview of the biological literature on modularity. 

Our work also relates to the study of graph clustering and network science. 
\citet{von2007tutorial} gives an overview of spectral clustering, the technique we use, as well as a large amount of work on the topic. Since it was published, \citet{lee2014multiway} derive an analogue of the Cheeger inequalities \cite{alon1985lambda1, dodziuk1984difference} for multi-way partitioning, providing a further justification for the spectral clustering algorithm.
\citet{barabasi2016network} give an introduction to the field of network science, which has produced studies such as those by \citet{girvan2002community} and \citet{newman2004finding} that study graph clustering under the name `community detection'.

A number of papers have investigated some aspect of the structure of neural networks. For instance, \citet{frankle2018the} discover that neural networks contain efficiently-trainable subnetworks, inspiring several follow-up papers \cite{zhou2019deconstructing, frankle2019linear}, and \citet{ramanujan2019whats} show that randomly-initialized neural networks contain subnetworks that achieve high performance on image classification problems. In general, the field of interpretability (sometimes known as transparency or explainable AI) investigates techniques to understand the functioning of neural networks. Books by \citet{molnar2019interpretable} and \citet{samek2019explainable} offer an overview of this research.
Most similar to our own work within the field of interpretability is \citet{cammarata2020thread}, which investigates `circuits', small groups of neurons that evaluate some intelligible function.
Also relevant are \citet{olah2017feature}, \citet{olah2018building}, and \citet{carter2019activation}, which visualize the features learned in neural networks without relying on detailed information about the data distribution.

\citet{molchanov2017variational} discuss how variational dropout can sparsify neural networks. Although this is related to our dropout results, it is important to note that their work relies on learned per-neuron dropout rates causing sparsity, while our work uses standard dropout, and finds an unusually high n-cut controlling for sparsity.

%% file: main_sections/conclusion.tex
In this paper, we showed that MLPs trained with weight pruning are often more modular than would be expected given only their overall connectivity and distribution of weights.
We also demonstrated that dropout significantly accentuates the modularity, and that the modularity is closely associated with learning.
Finally, we exhibited an exploratory analysis of the module structure of the networks. 

Several questions remain: Do these results extend to larger networks trained on larger datasets? What is the appropriate notion of modularity for convolutional neural networks? Why are networks so clusterable, and why does dropout have the effect that it does? Are clusters akin to functional regions of the human brain, and how can we tell? If modularity is desirable, can it be directly regularized for? We hope that follow-up research will shed light on these puzzles.

%% file: main_sections/broader_impact.tex
Since this research is fairly early-stage, we think that it is hard to speak precisely about who would benefit or be put at disadvantage from this research. We also do not think that it leverages biases in the data, rather it helps analyze a neural network which may itself leverage biases in the data.

We think that the most likely impact of this research outside academia, if it has any at all, is that clustering methods could be used to ensure that those deploying neural network systems have a greater understanding of the functioning of the system. This could make it possible to ameliorate aspects of the learned representations that are undesirable to those deploying it, for instance to ensure safety in safety-critical applications, or to remove behavior that is felt to reflect undesirable bias in the data. If these aspects cannot be fixed, those deploying the system could simply not deploy it, and use a different training method instead.

It is also possible that analyzing the representations learned by systems trained on some dataset could reveal insights about the dataset itself. We would expect this addition to humanity's knowledge to usually be broadly beneficial.

One potential negative impact of this work would be if it were to produce techniques that enabled a limited degree of transparency into neural networks, that were believed to enable a great degree of transparency. This could lead to an invalid sense of assurance of the safety of the system. As such, we think it is important to clearly emphasize that this paper provides almost no understanding of the specific function of `modules', and we are as yet unable to infer any aspects of the behavior of the network from the techniques described.

%% file: supplementary.tex
\newpage
\appendix

\numberwithin{table}{section}
\numberwithin{figure}{section}

\section{Supplementary Material}

\subsection{Probabilistic interpretation of n-cut}
\label{app:ncut_meaning}

\input{appendix_sections/ncut_meaning}

\subsection{Training details}
\label{app:train_hypers}

\input{appendix_sections/training_hyperparams}

\subsection{Choosing the number of clusters}
\label{app:n_clusters}

\input{appendix_sections/n_clusters}

\subsection{Clusterability data}
\label{app:clusterability_data}

\input{appendix_sections/clust_data}

\subsection{Mixture dataset results}
\label{app:mixture_datasets}

\input{appendix_sections/mixture_results}

\subsection{Lesion experiment data}
\label{app:lesion_experiment_data}

\input{appendix_sections/lesion_data}

%% file: appendix_sections/ncut_meaning.tex
As well as the formal definition given in subsection \ref{subsec:clustering_NNs_definitions}, n-cut has a more intuitive interpretation. Divide each edge between vertices $i$ and $j$ into two `stubs', one attached to $i$ and the other attached to $j$, and associate with each stub the weight of the whole edge. 

Now: suppose $(X_1, \dotsc, X_k)$ is a partition of the graph. First, pick an integer $l$ between 1 and $k$ uniformly at random. Secondly, out of all of the stubs attached to vertices in $X_l$, pick one with probability proportional to its weight. Say that this procedure `succeeds' if the edge associated with that stub connects two vertices inside $X_l$, and `fails' if the edge connects a vertex inside $X_l$ with a vertex outside $X_l$. The probability that the procedure fails is $\ncut(X_1, \dotsc, X_k)/k$.

Therefore, the n-cut divided by $k$ is roughly a measure of what proportion of edge weight coming from vertices inside a partition element crosses the partition boundary.

%% file: appendix_sections/training_hyperparams.tex
During training, we use the Adam algorithm \citep{kingma2014adam} with the standard Keras hyperparameters: learning rate $0.001$, $\beta_1 = 0.9$, $\beta_2 = 0.999$, no amsgrad. For pruning, our initial sparsity is 0.5, our final sparsity is 0.9, the pruning frequency is 10 steps, and we use a cubic pruning schedule (see \citet{zhu2017prune}). Initial and final sparsities were chosen due to their use in the TensorFlow Model Optimization Tutorial.\footnote{URL: \url{https://web.archive.org/web/20190817115045/https://www.tensorflow.org/model_optimization/guide/pruning/pruning_with_keras}} 
We use Tensorflow's implementation of the Keras API \cite{tensorflow2015-whitepaper, chollet2015keras}.

%% file: appendix_sections/n_clusters.tex
The number of clusters is a hyperparameter to the spectral clustering algorithm \cite{von2007tutorial}. In the paper we reported the results with using 4 and 10 clusters; the first for section~\ref{sec:clusterability} and the latter for section~\ref{sec:dependency_structure}. 

To test the robustness of our results to this hyperparameter, we re-ran our basic clusterability experiments three times where the number of clusters was 2, 7, and 10 respectively.
For 7 and 10 clusters, the results aligned with what we reported for 4 clusters in section~\ref{sec:clusterability}, including stronger significance patterns and lower mean n-cut for dropout-trained models vs. without dropout. Results are shown in tables \ref{tab:ncut-stats-7-clusters} and \ref{tab:ncut-stats-10-clusters} respectively.

\begin{table*}[tb]
\caption{
Clusterability results with 7 clusters. Reporting as in table~\ref{tab:ncut-stats-basic}.
}
\label{tab:ncut-stats-7-clusters}
\vskip 0.15in
\begin{center}
\begin{small}
\begin{tabular}{lccccccc}
\toprule
Dataset & Dropout & Prop.\ sig.\ & Train acc.\ & Test acc.\ &  N-cuts & Dist.\ n-cuts \\
\midrule
MNIST    & $\times$ & $1.0$ & $1.00 $ & $0.984$ & $4.556 \pm 0.052$ & $4.710 \pm 0.026$ \\
MNIST    & $\surd$ & $1.0$ & $0.967$ & $0.979$ & $4.222 \pm 0.035$ & $4.712 \pm 0.023$ \\
Fashion  & $\times$ & $1.0$ & $0.983$ & $0.893$ & $4.351 \pm 0.057$ & $4.613 \pm 0.029$ \\
Fashion  & $\surd$ & $1.0$ & $0.863$ & $0.869$ & $4.079 \pm 0.046$ & $4.663 \pm 0.029$ \\
CIFAR-10 & $\times$ & $0.1$ & $0.650$ & $0.415$ & $4.707  \pm 0.101$ & $4.647 \pm 0.024$ \\
CIFAR-10 & $\surd$ & $0.9$ & $0.427$ & $0.422$ & $4.373 \pm 0.160$ & $4.618 \pm 0.035$ \\
\bottomrule
\end{tabular}
\end{small}
\end{center}
\vskip -0.1in
\end{table*}

\begin{table*}[tb]
\caption{
Clusterability results with 10 clusters. Reporting as in table~\ref{tab:ncut-stats-basic}.
}
\label{tab:ncut-stats-10-clusters}
\vskip 0.15in
\begin{center}
\begin{small}
\begin{tabular}{lccccccc}
\toprule
Dataset & Dropout & Prop.\ sig.\ & Train acc.\ & Test acc.\ &  N-cuts & Dist.\ n-cuts \\
\midrule
MNIST    & $\times$ & $1.0$ & $1.00 $ & $0.984$ & $7.137 \pm 0.042$ & $7.397 \pm 0.037$ \\
MNIST    & $\surd$ & $1.0$ & $0.967$ & $0.979$ & $6.688 \pm 0.035$ & $7.421 \pm 0.031$ \\
Fashion  & $\times$ & $0.9$ & $0.983$ & $0.893$ & $6.975 \pm 0.117$ & $7.257 \pm 0.036$ \\
Fashion  & $\surd$ & $1.0$ & $0.863$ & $0.869$ & $6.460 \pm 0.041$ & $7.339 \pm 0.034$ \\
CIFAR-10 & $\times$ & $0.2$ & $0.650$ & $0.415$ & $7.331  \pm 0.125$ & $7.303 \pm 0.031$ \\
CIFAR-10 & $\surd$ & $1.0$ & $0.427$ & $0.422$ & $6.973 \pm 0.169$ & $7.298 \pm 0.048$ \\
\bottomrule
\end{tabular}
\end{small}
\end{center}
\vskip -0.1in
\end{table*}


When we ran the experiments with 2 clusters, the significance results were very different, as shown in table~\ref{tab:ncut-stats-2-clusters}. No network trained on MNIST or Fashion-MNIST was statistically significantly clusterable, but 9 out of 10 networks trained on CIFAR-10 were, whether or not dropout was used. As such, we conjecture that our results generalize to any number of clusters that is not too small.

\begin{table*}[tb]
\caption{
Clusterability results with 2 clusters. See table \ref{tab:ncut-stats-basic} for details.
}
\label{tab:ncut-stats-2-clusters}
\vskip 0.15in
\begin{center}
\begin{small}
\begin{tabular}{lccccccc}
\toprule
Dataset & Dropout & Prop.\ sig.\ & Train acc.\ & Test acc.\ &  N-cuts & Dist.\ n-cuts \\
\midrule
MNIST    & $\times$ & $0$ & $1.00 $ & $0.984$ & $0.333 \pm 0.003$ & $0.330 \pm 0.003$ \\
MNIST    & $\surd$ & $0$ & $0.967$ & $0.979$ & $0.332 \pm 0.002$ & $0.323 \pm 0.003$ \\
Fashion  & $\times$ & $0$ & $0.983$ & $0.893$ & $0.319 \pm 0.003$ & $0.313 \pm 0.003$ \\
Fashion  & $\surd$ & $0$ & $0.863$ & $0.869$ & $0.312 \pm 0.003$ & $0.312 \pm 0.003$ \\
CIFAR-10 & $\times$ & $0.9$ & $0.650$ & $0.415$ & $0.306  \pm 0.010$ & $0.313 \pm 0.006$ \\
CIFAR-10 & $\surd$ & $0.9$ & $0.427$ & $0.422$ & $0.290 \pm 0.015$ & $0.308 \pm 0.008$ \\
\bottomrule
\end{tabular}
\end{small}
\end{center}
\vskip -0.1in
\end{table*}


%% file: appendix_sections/clust_data.tex
Tables~\ref{tab:mnist-unpruned-no-dropout}, \ref{tab:mnist-unpruned-dropout}, \ref{tab:mnist-pruned-no-dropout}, \ref{tab:mnist-pruned-dropout}, \ref{tab:fashion-unpruned-no-dropout}, \ref{tab:fashion-unpruned-dropout}, \ref{tab:fashion-pruned-no-dropout}, \ref{tab:fashion-pruned-dropout}, \ref{tab:cifar-unpruned-no-dropout}, \ref{tab:cifar-unpruned-dropout}, \ref{tab:cifar-pruned-no-dropout}, and~\ref{tab:cifar-pruned-dropout} show detailed information about clusterability of each network trained on MNIST, Fashion-MNIST, and CIFAR-10, with and without dropout, both before pruning was applied and at the end of training. They include statistics for how each network compares to the distribution of shuffles of that network's weights, as well as the distribution of shuffles of the non-zero weights that preserve the network topology. They also include the Cohen's $d$ statistic of the normalized difference between the mean of the standard shuffle distribution and the mean of the non-zero shuffle distribution for each trained network. If $\mu_1$ and $\mu_2$ are the respective sample means, $\sigma_1$ and $\sigma_2$ are the respective sample standard deviations, and $n_1$ and $n_2$ are the respective sample sizes, then Cohen's $d$ is defined as
\begin{equation*}
 d = (\mu_1 - \mu_2)/\sigma_\textrm{pooled},
\end{equation*}
where
\begin{equation*}
    \sigma_\textrm{pooled} = \sqrt{\frac{(n_1 - 1) \sigma_1^2 + (n_2 - 1) \sigma_2^2}{n_1 + n_2 - 2}}.
\end{equation*}

Note that different samples were drawn for the generation of this table than were used to calculate the statistics in the main text, so there may be minor discrepancies. The run ID of networks is the same before and after pruning, so each row in a table of networks after pruning refers to the pruned version of the network whose information is displayed in the corresponding row of the corresponding table of networks before pruning.

\begin{figure}[htb]
\vskip 0.2in
\begin{center}
\centerline{\includegraphics[width=0.7\textwidth]{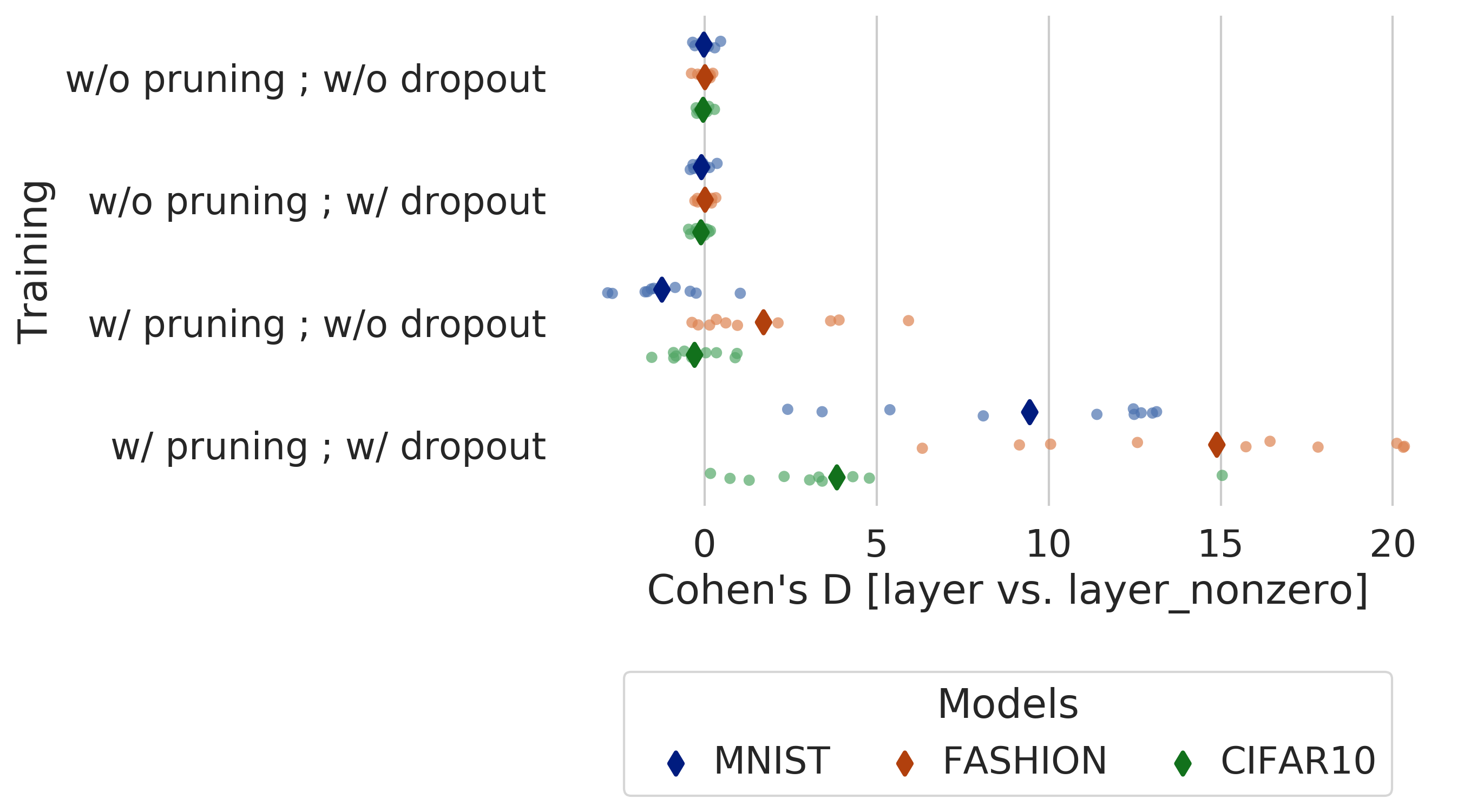}}
\caption{Cohen's $d$ statistics comparing the distribution of n-cuts of per-layer shuffles of weights (denoted `layer' in the figure) to the distribution of n-cuts of per-layer shuffles of only non-zero weights (denoted `layer\_nonzero' in the figure). For each method of training, we show results for MNIST, then Fashion-MNIST, then CIFAR-10. Positive numbers mean that the `layer' distribution had higher n-cuts. Each individual light circle is the Cohen's $d$ for the shuffle distributions of a single trained network, and the bolded diamonds are the means of the Cohen's $d$s.}
\label{fig:cohens-d}
\end{center}
\vskip -0.2in
\end{figure}

In figure~\ref{fig:cohens-d}, we graphically show the Cohen's $d$ statistics documented in tables~\ref{tab:mnist-unpruned-no-dropout}, \ref{tab:mnist-unpruned-dropout}, \ref{tab:mnist-pruned-no-dropout}, \ref{tab:mnist-pruned-dropout}, \ref{tab:fashion-unpruned-no-dropout}, \ref{tab:fashion-unpruned-dropout}, \ref{tab:fashion-pruned-no-dropout}, \ref{tab:fashion-pruned-dropout}, \ref{tab:cifar-unpruned-no-dropout}, \ref{tab:cifar-unpruned-dropout}, \ref{tab:cifar-pruned-no-dropout}, and~\ref{tab:cifar-pruned-dropout}. Since networks should have very few weights that are exactly zero before pruning, we expect the $d$ statistics to be approximately zero for unpruned networks, and indeed this is what we see. Note the negative average $d$ statistic for networks trained on MNIST without dropout, indicating that the actual trained topologies are less clusterable on average than random shuffles of the networks.

\begin{figure}[htb]
\vskip 0.2in
\begin{center}
\centerline{\includegraphics[width=\textwidth]{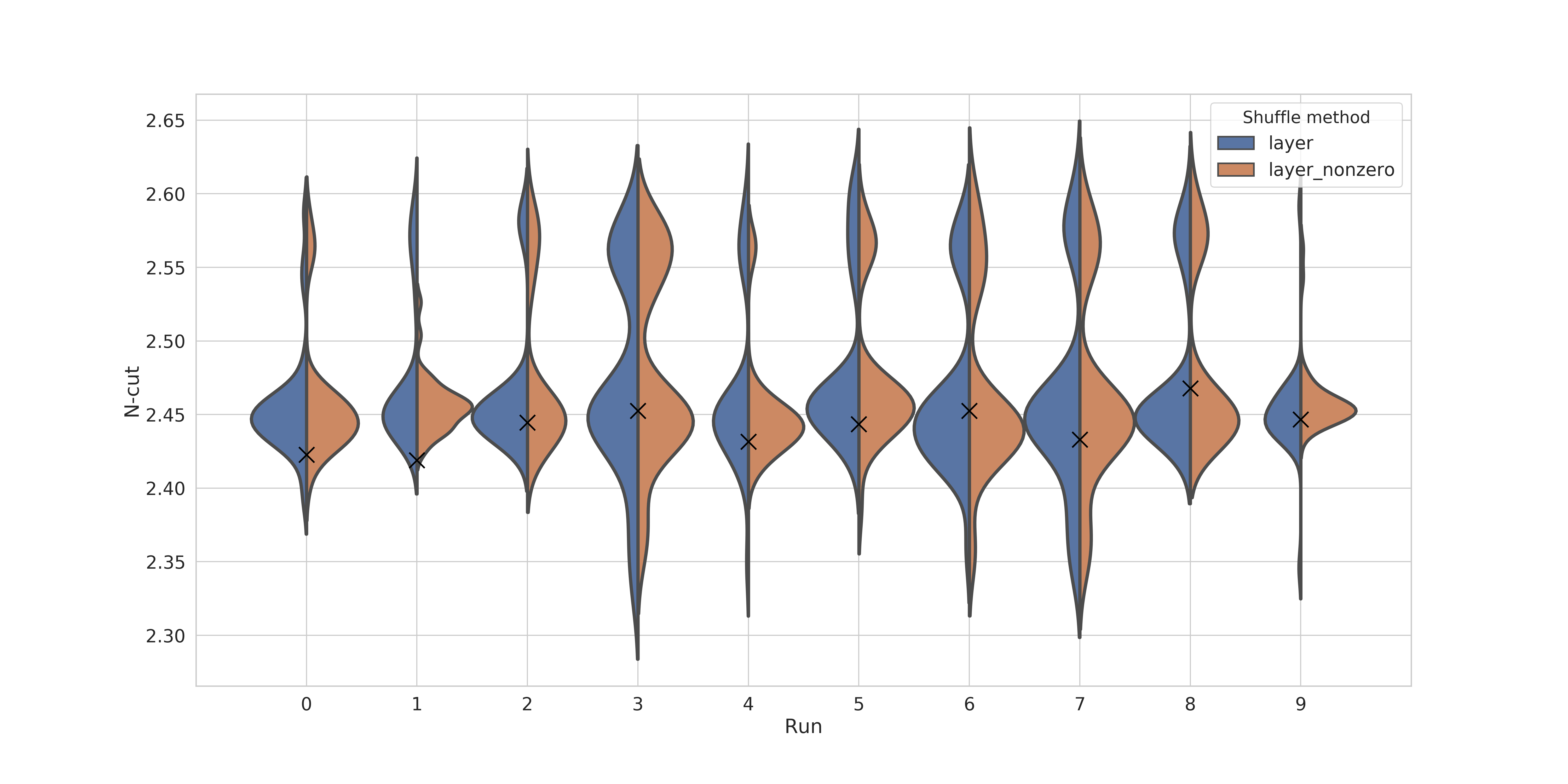}}
\caption{Crosses mark actual n-cuts of networks trained on MNIST \textbf{without dropout, before pruning was applied}. The blue distribution (on the left of each central line) is of n-cuts of shuffles of weights of the trained network, and the orange distribution (on the right of each central line) is of n-cuts of shuffles of the non-zero weights of the trained network. Plot produced by the \texttt{violinplot} function of seaborn 0.9.0 \cite{seaborn_0_9_0} with default bandwidth.}
\label{fig:violin-unpruned-no-dropout}
\end{center}
\vskip -0.2in
\end{figure}

\begin{figure}[htb]
\vskip 0.2in
\begin{center}
\centerline{\includegraphics[width=\textwidth]{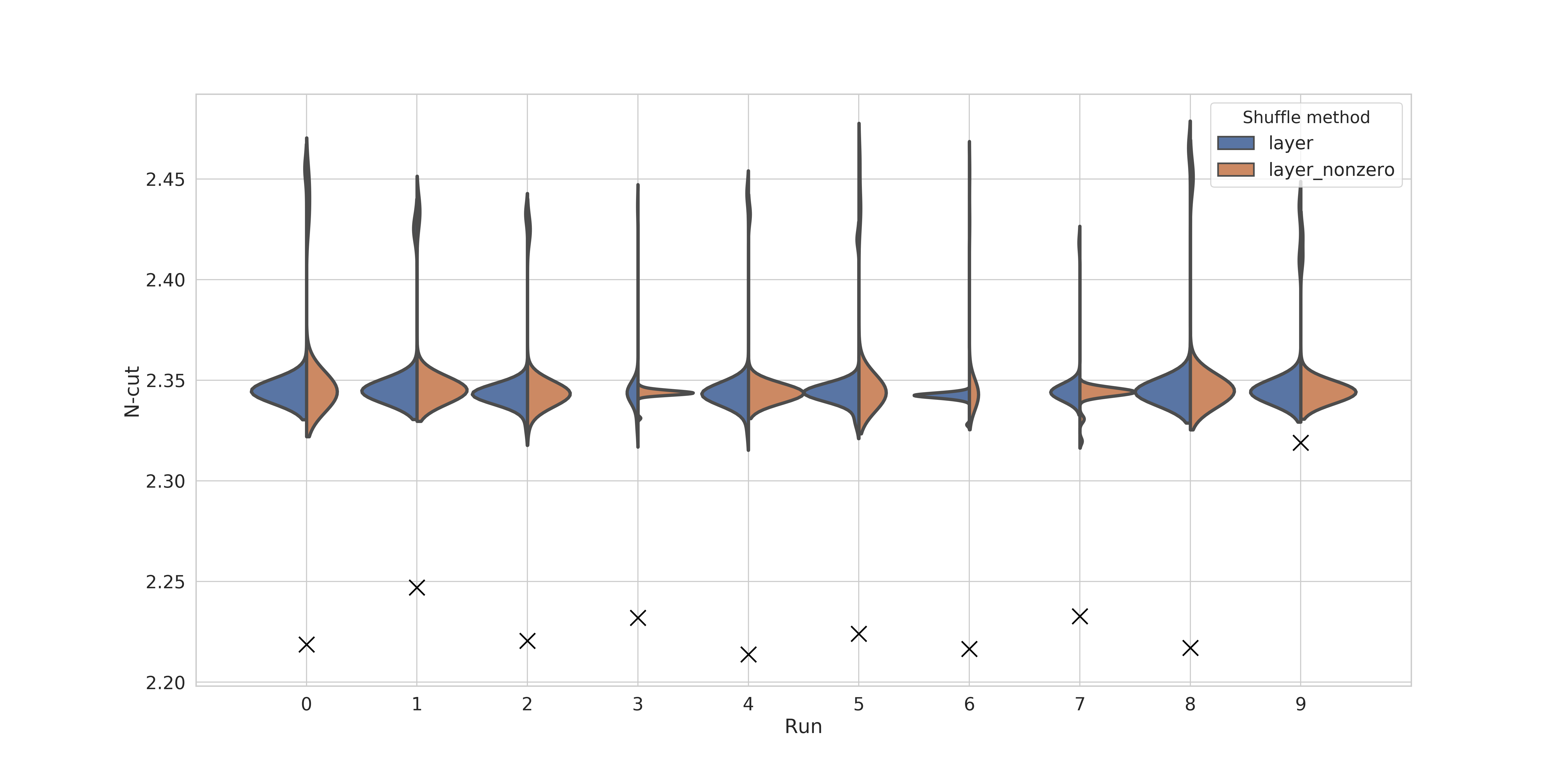}}
\caption{Crosses mark actual n-cuts of networks trained on MNIST \textbf{with dropout, before pruning was applied}. The blue distribution (on the left of each central line) is of n-cuts of shuffles of weights of the trained network, and the orange distribution (on the right of each central line) is of n-cuts of shuffles of the non-zero weights of the trained network. Plot produced by the \texttt{violinplot} function of seaborn 0.9.0 \cite{seaborn_0_9_0} with default bandwidth.}
\label{fig:violin-unpruned-dropout}
\end{center}
\vskip -0.2in
\end{figure}

\begin{figure}[htb]
\vskip 0.2in
\begin{center}
\centerline{\includegraphics[width=\textwidth]{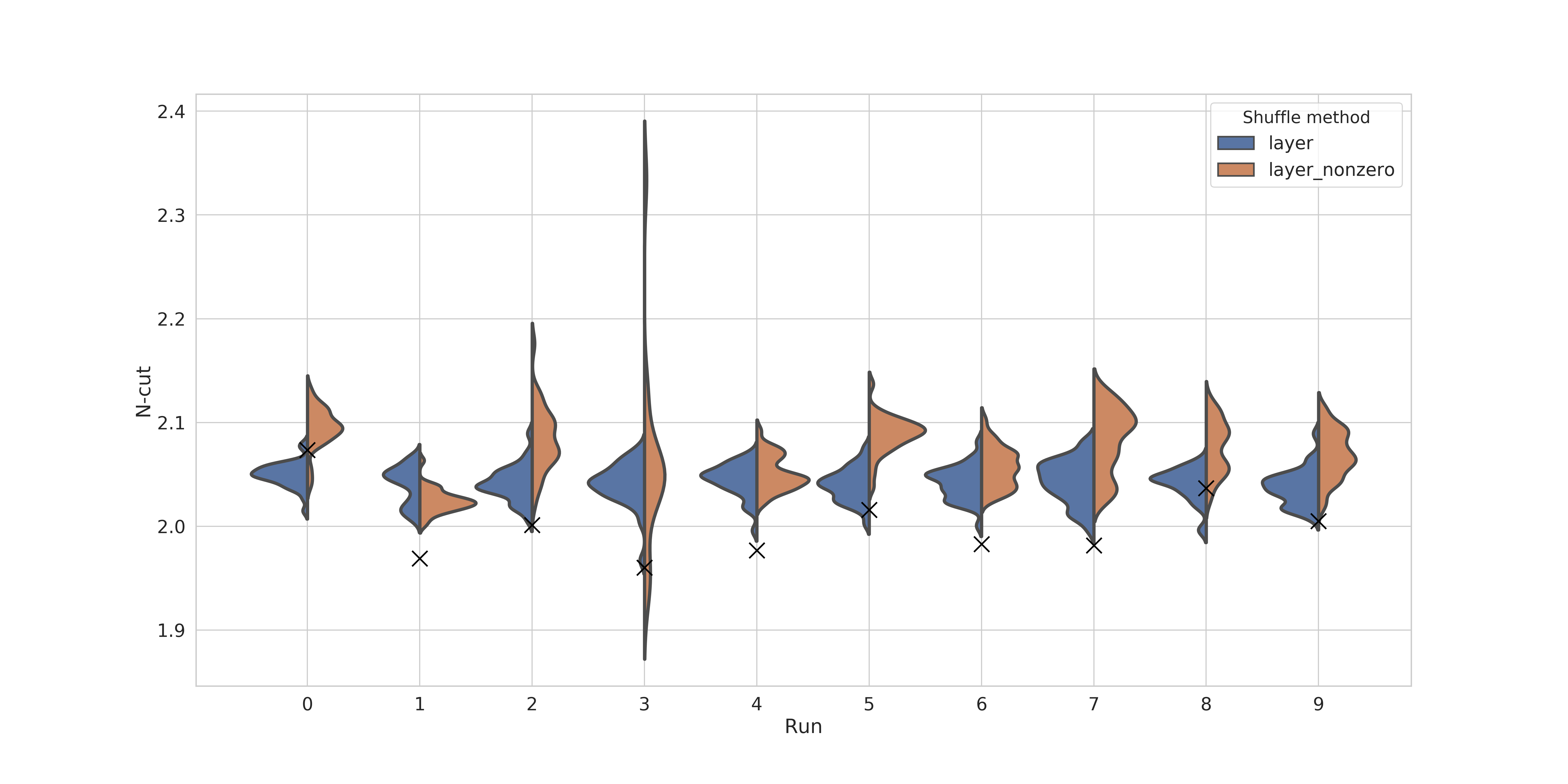}}
\caption{Crosses mark actual n-cuts of networks trained on MNIST \textbf{without dropout at the end of training}. The blue distribution (on the left of each central line) is of n-cuts of shuffles of weights of the trained network, and the orange distribution (on the right of each central line) is of n-cuts of shuffles of the non-zero weights of the trained network. Plot produced by the \texttt{violinplot} function of seaborn 0.9.0 \cite{seaborn_0_9_0} with default bandwidth.}
\label{fig:violin-pruned-no-dropout}
\end{center}
\vskip -0.2in
\end{figure}

\begin{figure}[htb]
\vskip 0.2in
\begin{center}
\centerline{\includegraphics[width=\textwidth]{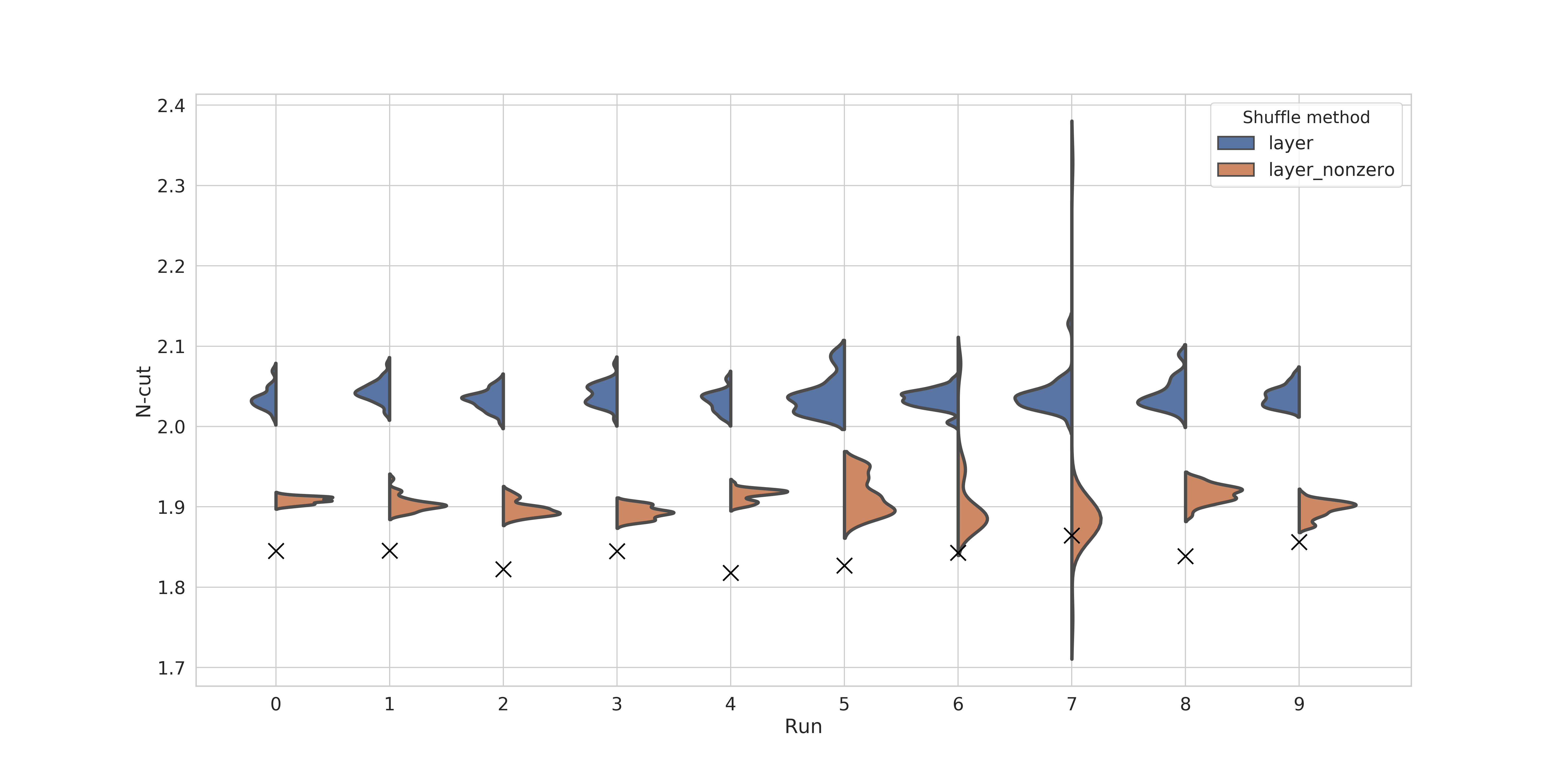}}
\caption{Crosses mark actual n-cuts of networks trained on MNIST \textbf{with dropout at the end of training}. The blue distribution (on the left of each central line) is of n-cuts of shuffles of weights of the trained network, and the orange distribution (on the right of each central line) is of n-cuts of shuffles of the non-zero weights of the trained network. Plot produced by the \texttt{violinplot} function of seaborn 0.9.0 \cite{seaborn_0_9_0} with default bandwidth.}
\label{fig:violin-pruned-dropout}
\end{center}
\vskip -0.2in
\end{figure}

The distributions of n-cuts for shuffles of networks trained on MNIST, as well as the distributions of n-cuts for shuffles of the non-zero elements, as well as the n-cut of the actual network, is shown in a series of violin plots. Figure~\ref{fig:violin-unpruned-no-dropout} shows the distributions for networks trained without dropout before the onset of pruning, figure~\ref{fig:violin-unpruned-dropout} shows the distributions for networks trained with dropout before the onset of pruning, figure~\ref{fig:violin-pruned-no-dropout} shows the distributions for networks trained without dropout at the end of training, and figure~\ref{fig:violin-pruned-dropout} shows the distributions for networks trained with dropout at the end of training. Note that since networks have almost no weights that are exactly 0 before pruning, the blue and orange distributions in figures~\ref{fig:violin-unpruned-no-dropout} and \ref{fig:violin-unpruned-dropout} are nearly identical. Note also that the distributions for shuffles of unpruned networks trained without dropout tend to be bimodal, and the distributions for shuffles of unpruned networks trained with dropout tend to have long right tails. We do not have an explanation for either of these phenomena.

\begin{table}[htb]
\caption{Information about each network trained on \textbf{MNIST without} dropout \textbf{before} pruning. $p$-values are calculated using the method described in \citet{north2002note}. ``Dist.\ n-cuts'' shows the mean and standard deviation of the distribution of n-cuts of shuffled networks. We also present statistics for the distribution of n-cuts of networks generated by shuffling only the non-zero elements of the trained network, leaving the topology intact, as described in subsection~\ref{subsec:connectivity_structure}. ``$p$-value (non-zero)'' refers to the $p$-value with respect to this distribution, and ``Non-zero dist.\ n-cuts'' shows the mean and standard deviation of this distribution. We draw 320 samples per run from each distribution. Cohen's $d$ is the normalized difference between the mean of the standard shuffle distribution and the mean of the non-zero shuffle distribution. $p$-values less than $1/320$ are bolded.}
\label{tab:mnist-unpruned-no-dropout}
\vskip 0.15in
\begin{center}
\begin{small}
\begin{tabular}{lcccccccc}
\toprule
Run & N-cut   & $p$-value        & Dist.\ n-cuts     & $p$-value (non-zero) & Non-zero dist.\ n-cuts & Cohen's $d$ \\
\midrule
0   & $2.423$ & $0.065$          & $2.456 \pm 0.038$ & $0.065$              & $2.459 \pm 0.043$      & $-0.087$      \\
1   & $2.419$ & $\mathbf{0.003}$ & $2.474 \pm 0.050$ & $\mathbf{0.003}$     & $2.456 \pm 0.020$      & $+0.468 $      \\
2   & $2.445$ & $0.346$          & $2.465 \pm 0.045$ & $0.346$              & $2.479 \pm 0.056$      & $-0.280$      \\
3   & $2.453$ & $0.439$          & $2.474 \pm 0.069$ & $0.502$              & $2.481 \pm 0.066$      & $-0.104$      \\
4   & $2.432$ & $0.190$          & $2.467 \pm 0.058$ & $0.190$              & $2.452 \pm 0.038$      & $+0.297 $      \\
5   & $2.443$ & $0.159$          & $2.483 \pm 0.056$ & $0.190$              & $2.477 \pm 0.051$      & $+0.115 $      \\
6   & $2.453$ & $0.595$          & $2.465 \pm 0.058$ & $0.657$              & $2.469 \pm 0.064$      & $-0.067$      \\
7   & $2.433$ & $0.252$          & $2.459 \pm 0.067$ & $0.190$              & $2.465 \pm 0.066$      & $-0.090$      \\
8   & $2.468$ & $0.751$          & $2.479 \pm 0.055$ & $0.657$              & $2.483 \pm 0.059$      & $-0.089$      \\
9   & $2.447$ & $0.502$          & $2.450 \pm 0.033$ & $0.097$              & $2.461 \pm 0.025$      & $-0.344$      \\
\bottomrule
\end{tabular}
\end{small}
\end{center}
\vskip -0.1in
\end{table}

\begin{table}[htb]
\caption{Information about each network trained on \textbf{MNIST with} dropout \textbf{before} pruning. All information as in table~\ref{tab:mnist-unpruned-no-dropout}. $p$-values less than $1/320$ are bolded.}
\label{tab:mnist-unpruned-dropout}
\vskip 0.15in
\begin{center}
\begin{small}
\begin{tabular}{lcccccccc}
\toprule
Run & N-cut   & $p$-value        & Dist.\ n-cuts     & $p$-value (non-zero) & Non-zero dist.\ n-cuts & Cohen's $d$ \\
\midrule
0   & $2.219$ & $\mathbf{0.003}$ & $2.348 \pm 0.019$ & $\mathbf{0.003}$     & $2.356 \pm 0.032$      & $-0.310$ \\
1   & $2.247$ & $\mathbf{0.003}$ & $2.350 \pm 0.020$ & $\mathbf{0.003}$     & $2.351 \pm 0.022$      & $-0.047$ \\
2   & $2.221$ & $\mathbf{0.003}$ & $2.345 \pm 0.016$ & $\mathbf{0.003}$     & $2.348 \pm 0.020$      & $-0.142$ \\
3   & $2.232$ & $\mathbf{0.003}$ & $2.345 \pm 0.017$ & $\mathbf{0.003}$     & $2.344 \pm 0.002$      & $+0.145$ \\
4   & $2.214$ & $\mathbf{0.003}$ & $2.346 \pm 0.018$ & $\mathbf{0.003}$     & $2.346 \pm 0.016$      & $-0.035$ \\
5   & $2.224$ & $\mathbf{0.003}$ & $2.345 \pm 0.014$ & $\mathbf{0.003}$     & $2.353 \pm 0.029$      & $-0.334$ \\
6   & $2.216$ & $\mathbf{0.003}$ & $2.342 \pm 0.003$ & $\mathbf{0.003}$     & $2.349 \pm 0.024$      & $-0.414$ \\
7   & $2.233$ & $\mathbf{0.003}$ & $2.346 \pm 0.013$ & $\mathbf{0.003}$     & $2.343 \pm 0.005$      & $+0.365$ \\
8   & $2.217$ & $\mathbf{0.003}$ & $2.348 \pm 0.021$ & $\mathbf{0.003}$     & $2.351 \pm 0.026$      & $-0.145$ \\
9   & $2.319$ & $\mathbf{0.003}$ & $2.349 \pm 0.019$ & $\mathbf{0.003}$     & $2.349 \pm 0.018$      & $+0.027$ \\
\bottomrule
\end{tabular}
\end{small}
\end{center}
\vskip -0.1in
\end{table}

\begin{table}[htb]
\caption{Information about each network trained on \textbf{MNIST without} dropout \textbf{after} pruning. All information as in table~\ref{tab:mnist-unpruned-no-dropout}. Note that each row corresponds to the network in the same row in table~\ref{tab:mnist-unpruned-no-dropout} after undergoing pruning. $p$-values less than $1/320$ are bolded.}
\label{tab:mnist-pruned-no-dropout}
\vskip 0.15in
\begin{center}
\begin{small}
\begin{tabular}{lcccccccc}
\toprule
Run & N-cut   & $p$-value        & Dist.\ n-cuts     & $p$-value (non-zero) & Non-zero dist.\ n-cuts & Cohen's $d$ \\
\midrule
0   & $2.073$ & $0.938$          & $2.050 \pm 0.012$ & $0.097$              & $2.094 \pm 0.020$      & $-2.678$ \\
1   & $1.969$ & $\mathbf{0.003}$ & $2.040 \pm 0.018$ & $\mathbf{0.003}$     & $2.024 \pm 0.012$      & $+1.039$ \\
2   & $2.001$ & $\mathbf{0.003}$ & $2.041 \pm 0.017$ & $\mathbf{0.003}$     & $2.084 \pm 0.031$      & $-1.726$ \\
3   & $1.960$ & $\mathbf{0.003}$ & $2.041 \pm 0.021$ & $0.128$              & $2.056 \pm 0.087$      & $-0.243$ \\
4   & $1.977$ & $\mathbf{0.003}$ & $2.044 \pm 0.016$ & $\mathbf{0.003}$     & $2.051 \pm 0.017$      & $-0.417$ \\
5   & $2.016$ & $0.065$          & $2.040 \pm 0.016$ & $\mathbf{0.003}$     & $2.088 \pm 0.018$      & $-2.812$ \\
6   & $1.983$ & $\mathbf{0.003}$ & $2.041 \pm 0.016$ & $\mathbf{0.003}$     & $2.057 \pm 0.019$      & $-0.853$ \\
7   & $1.982$ & $\mathbf{0.003}$ & $2.043 \pm 0.021$ & $\mathbf{0.003}$     & $2.083 \pm 0.032$      & $-1.473$ \\
8   & $2.037$ & $0.346$          & $2.040 \pm 0.016$ & $0.065$              & $2.076 \pm 0.026$      & $-1.659$ \\
9   & $2.005$ & $\mathbf{0.003}$ & $2.037 \pm 0.017$ & $\mathbf{0.003}$     & $2.069 \pm 0.023$      & $-1.547$ \\
\bottomrule
\end{tabular}
\end{small}
\end{center}
\vskip -0.1in
\end{table}

\begin{table}[htb]
\caption{Information about each network trained on \textbf{MNIST with} dropout \textbf{after} pruning. All information as in table~\ref{tab:mnist-unpruned-no-dropout}. Note that each row corresponds to the network in the same row in table~\ref{tab:mnist-unpruned-dropout} after undergoing pruning. $p$-values less than $1/320$ are bolded.}
\label{tab:mnist-pruned-dropout}
\vskip 0.15in
\begin{center}
\begin{small}
\begin{tabular}{lcccccccc}
\toprule
Run & N-cut   & $p$-value        & Dist.\ n-cuts     & $p$-value (non-zero) & Non-zero dist.\ n-cuts & Cohen's $d$ \\
\midrule
0   & $1.845$ & $\mathbf{0.003}$ & $2.035 \pm 0.013$ & $\mathbf{0.003}$     & $1.908 \pm 0.004$      & $+13.015$      \\
1   & $1.846$ & $\mathbf{0.003}$ & $2.044 \pm 0.013$ & $\mathbf{0.003}$     & $1.904 \pm 0.009$      & $+12.461$      \\
2   & $1.822$ & $\mathbf{0.003}$ & $2.033 \pm 0.012$ & $\mathbf{0.003}$     & $1.897 \pm 0.009$      & $+12.687$      \\
3   & $1.845$ & $\mathbf{0.003}$ & $2.040 \pm 0.014$ & $\mathbf{0.003}$     & $1.893 \pm 0.008$      & $+13.135$      \\
4   & $1.818$ & $\mathbf{0.003}$ & $2.032 \pm 0.012$ & $\mathbf{0.003}$     & $1.914 \pm 0.008$      & $+11.401$      \\
5   & $1.827$ & $\mathbf{0.003}$ & $2.038 \pm 0.023$ & $\mathbf{0.003}$     & $1.914 \pm 0.023$      & $+5.387 $      \\
6   & $1.843$ & $\mathbf{0.003}$ & $2.034 \pm 0.012$ & $\mathbf{0.003}$     & $1.909 \pm 0.050$      & $+3.417 $      \\
7   & $1.864$ & $\mathbf{0.003}$ & $2.038 \pm 0.020$ & $0.034$              & $1.895 \pm 0.081$      & $+2.417 $      \\
8   & $1.839$ & $\mathbf{0.003}$ & $2.040 \pm 0.018$ & $\mathbf{0.003}$     & $1.916 \pm 0.011$      & $+8.101 $      \\
9   & $1.856$ & $\mathbf{0.003}$ & $2.037 \pm 0.012$ & $\mathbf{0.003}$     & $1.897 \pm 0.010$      & $+12.485$      \\
\bottomrule
\end{tabular}
\end{small}
\end{center}
\vskip -0.1in
\end{table}

\begin{table}[htb]
\caption{Information about each network trained on \textbf{Fashion-MNIST without} dropout \textbf{before} pruning. All information as in table~\ref{tab:mnist-unpruned-no-dropout}. $p$-values less than $1/320$ are bolded.}
\label{tab:fashion-unpruned-no-dropout}
\vskip 0.15in
\begin{center}
\begin{small}
\begin{tabular}{lcccccccc}
\toprule
Run & N-cut   & $p$-value        & Dist.\ n-cuts     & $p$-value (non-zero) & Non-zero dist.\ n-cuts & Cohen's $d$ \\
\midrule
0   & $2.185$ & $\mathbf{0.003}$ & $2.335 \pm 0.027$ & $\mathbf{0.003}$     & $2.330 \pm 0.021$      & $+0.242$ \\
1   & $2.167$ & $\mathbf{0.003}$ & $2.324 \pm 0.028$ & $\mathbf{0.003}$     & $2.320 \pm 0.024$      & $+0.163$ \\
2   & $2.171$ & $\mathbf{0.003}$ & $2.321 \pm 0.020$ & $\mathbf{0.003}$     & $2.324 \pm 0.047$      & $-0.094$ \\
3   & $2.178$ & $\mathbf{0.003}$ & $2.321 \pm 0.020$ & $\mathbf{0.003}$     & $2.318 \pm 0.020$      & $+0.171$ \\
4   & $2.182$ & $\mathbf{0.003}$ & $2.276 \pm 0.026$ & $\mathbf{0.003}$     & $2.281 \pm 0.026$      & $-0.199$ \\
5   & $2.183$ & $\mathbf{0.003}$ & $2.322 \pm 0.020$ & $\mathbf{0.003}$     & $2.319 \pm 0.024$      & $+0.130$ \\
6   & $2.179$ & $\mathbf{0.003}$ & $2.325 \pm 0.022$ & $\mathbf{0.003}$     & $2.338 \pm 0.042$      & $-0.379$ \\
7   & $2.189$ & $\mathbf{0.003}$ & $2.333 \pm 0.016$ & $\mathbf{0.003}$     & $2.334 \pm 0.022$      & $-0.012$ \\
8   & $2.180$ & $\mathbf{0.003}$ & $2.315 \pm 0.024$ & $\mathbf{0.003}$     & $2.312 \pm 0.026$      & $+0.134$ \\
9   & $2.172$ & $\mathbf{0.003}$ & $2.333 \pm 0.020$ & $\mathbf{0.003}$     & $2.333 \pm 0.020$      & $-0.027$ \\
\bottomrule
\end{tabular}
\end{small}
\end{center}
\vskip -0.1in
\end{table}

\begin{table}[htb]
\caption{Information about each network trained on \textbf{Fashion-MNIST with} dropout \textbf{before} pruning. All information as in table~\ref{tab:mnist-unpruned-no-dropout}. $p$-values less than $1/320$ are bolded.}
\label{tab:fashion-unpruned-dropout}
\vskip 0.15in
\begin{center}
\begin{small}
\begin{tabular}{lcccccccc}
\toprule
Run & N-cut   & $p$-value        & Dist.\ n-cuts     & $p$-value (non-zero) & Non-zero dist.\ n-cuts & Cohen's $d$ \\
\midrule
0   & $2.140$ & $\mathbf{0.003}$ & $2.313 \pm 0.026$ & $\mathbf{0.003}$     & $2.304 \pm 0.024$      & $+0.329$ \\
1   & $2.155$ & $\mathbf{0.003}$ & $2.356 \pm 0.045$ & $\mathbf{0.003}$     & $2.357 \pm 0.041$      & $-0.014$ \\
2   & $2.143$ & $\mathbf{0.003}$ & $2.308 \pm 0.031$ & $\mathbf{0.003}$     & $2.316 \pm 0.026$      & $-0.280$ \\
3   & $2.155$ & $\mathbf{0.003}$ & $2.350 \pm 0.038$ & $\mathbf{0.003}$     & $2.359 \pm 0.053$      & $-0.196$ \\
4   & $2.147$ & $\mathbf{0.003}$ & $2.335 \pm 0.055$ & $\mathbf{0.003}$     & $2.325 \pm 0.026$      & $+0.215$ \\
5   & $2.147$ & $\mathbf{0.003}$ & $2.329 \pm 0.020$ & $\mathbf{0.003}$     & $2.326 \pm 0.030$      & $+0.113$ \\
6   & $2.136$ & $\mathbf{0.003}$ & $2.292 \pm 0.033$ & $\mathbf{0.003}$     & $2.291 \pm 0.028$      & $+0.032$ \\
7   & $2.148$ & $\mathbf{0.003}$ & $2.322 \pm 0.033$ & $\mathbf{0.003}$     & $2.323 \pm 0.045$      & $-0.029$ \\
8   & $2.147$ & $\mathbf{0.003}$ & $2.303 \pm 0.032$ & $\mathbf{0.003}$     & $2.309 \pm 0.025$      & $-0.207$ \\
9   & $2.132$ & $\mathbf{0.003}$ & $2.318 \pm 0.029$ & $\mathbf{0.003}$     & $2.312 \pm 0.030$      & $+0.202$ \\
\bottomrule
\end{tabular}
\end{small}
\end{center}
\vskip -0.1in
\end{table}

\begin{table}[htb]
\caption{Information about each network trained on \textbf{Fashion-MNIST without} dropout \textbf{after} pruning. All information as in table~\ref{tab:mnist-unpruned-no-dropout}. Note that each row corresponds to the network in the same row in table~\ref{tab:fashion-unpruned-no-dropout} after undergoing pruning. $p$-values less than $1/320$ are bolded.}
\label{tab:fashion-pruned-no-dropout}
\vskip 0.15in
\begin{center}
\begin{small}
\begin{tabular}{lcccccccc}
\toprule
Run & N-cut   & $p$-value        & Dist.\ n-cuts     & $p$-value (non-zero) & Non-zero dist.\ n-cuts & Cohen's $d$ \\
\midrule
0   & $1.876$ & $\mathbf{0.003}$ & $2.005 \pm 0.013$ & $0.034$              & $1.994 \pm 0.103$      & $+0.149$ \\
1   & $1.871$ & $\mathbf{0.003}$ & $1.992 \pm 0.014$ & $0.065$              & $1.942 \pm 0.072$      & $+0.958$ \\
2   & $1.839$ & $\mathbf{0.003}$ & $1.986 \pm 0.018$ & $0.034$              & $1.870 \pm 0.021$      & $+5.928$ \\
3   & $1.852$ & $\mathbf{0.003}$ & $1.992 \pm 0.021$ & $0.034$              & $2.011 \pm 0.139$      & $-0.184$ \\
4   & $1.909$ & $\mathbf{0.003}$ & $1.987 \pm 0.016$ & $0.034$              & $1.963 \pm 0.094$      & $+0.344$ \\
5   & $1.868$ & $\mathbf{0.003}$ & $1.989 \pm 0.018$ & $0.128$              & $1.892 \pm 0.033$      & $+3.664$ \\
6   & $1.936$ & $\mathbf{0.003}$ & $1.986 \pm 0.017$ & $0.159$              & $2.016 \pm 0.116$      & $-0.363$ \\
7   & $1.912$ & $\mathbf{0.003}$ & $2.003 \pm 0.021$ & $0.065$              & $1.931 \pm 0.015$      & $+3.910$ \\
8   & $1.870$ & $\mathbf{0.003}$ & $1.990 \pm 0.013$ & $0.221$              & $1.877 \pm 0.073$      & $+2.137$ \\
9   & $1.865$ & $\mathbf{0.003}$ & $1.997 \pm 0.019$ & $0.252$              & $1.930 \pm 0.151$      & $+0.619$ \\
\bottomrule
\end{tabular}
\end{small}
\end{center}
\vskip -0.1in
\end{table}

\begin{table}[htb]
\caption{Information about each network trained on \textbf{Fashion-MNIST with} dropout \textbf{after} pruning. All information as in table~\ref{tab:mnist-unpruned-no-dropout}. Note that each row corresponds to the network in the same row in table~\ref{tab:fashion-unpruned-dropout} after undergoing pruning. $p$-values less than $1/320$ are bolded.}
\label{tab:fashion-pruned-dropout}
\vskip 0.15in
\begin{center}
\begin{small}
\begin{tabular}{lcccccccc}
\toprule
Run & N-cut   & $p$-value        & Dist.\ n-cuts     & $p$-value (non-zero) & Non-zero dist.\ n-cuts & Cohen's $d$ \\
\midrule
0   & $1.747$ & $\mathbf{0.003}$ & $2.018 \pm 0.016$ & $\mathbf{0.003}$     & $1.822 \pm 0.008$      & $+15.732$ \\
1   & $1.709$ & $\mathbf{0.003}$ & $2.015 \pm 0.015$ & $\mathbf{0.003}$     & $1.779 \pm 0.007$      & $+20.336$ \\
2   & $1.722$ & $\mathbf{0.003}$ & $2.017 \pm 0.018$ & $0.034$              & $1.808 \pm 0.023$      & $+10.057$ \\
3   & $1.744$ & $\mathbf{0.003}$ & $2.014 \pm 0.016$ & $0.097$              & $1.806 \pm 0.044$      & $+6.330$  \\
4   & $1.730$ & $\mathbf{0.003}$ & $2.015 \pm 0.018$ & $\mathbf{0.003}$     & $1.783 \pm 0.008$      & $+16.432$ \\
5   & $1.682$ & $\mathbf{0.003}$ & $2.023 \pm 0.017$ & $\mathbf{0.003}$     & $1.762 \pm 0.005$      & $+20.306$ \\
6   & $1.713$ & $\mathbf{0.003}$ & $2.014 \pm 0.019$ & $\mathbf{0.003}$     & $1.791 \pm 0.016$      & $+12.580$ \\
7   & $1.751$ & $\mathbf{0.003}$ & $2.016 \pm 0.014$ & $\mathbf{0.003}$     & $1.803 \pm 0.006$      & $+20.116$ \\
8   & $1.746$ & $\mathbf{0.003}$ & $2.008 \pm 0.018$ & $0.034$              & $1.800 \pm 0.027$      & $+9.150$  \\
9   & $1.721$ & $\mathbf{0.003}$ & $2.010 \pm 0.015$ & $\mathbf{0.003}$     & $1.784 \pm 0.009$      & $+17.827$ \\
\bottomrule
\end{tabular}
\end{small}
\end{center}
\vskip -0.1in
\end{table}

\begin{table}[htb]
\caption{Information about each network trained on \textbf{CIFAR-10 without} dropout \textbf{before} pruning. All information as in table~\ref{tab:mnist-unpruned-no-dropout}. $p$-values less than $1/320$ and equal to $1$ are bolded.}
\label{tab:cifar-unpruned-no-dropout}
\vskip 0.15in
\begin{center}
\begin{small}
\begin{tabular}{lcccccccc}
\toprule
Run & N-cut   & $p$-value        & Dist.\ n-cuts     & $p$-value (non-zero) & Non-zero dist.\ n-cuts & Cohen's $d$ \\
\midrule
0   & $2.405$ & $\mathbf{1.000}$ & $2.247 \pm 0.013$ & $\mathbf{1.000}$     & $2.249 \pm 0.018$      & $-0.118$ \\
1   & $2.433$ & $\mathbf{1.000}$ & $2.249 \pm 0.014$ & $\mathbf{1.000}$     & $2.251 \pm 0.015$      & $-0.077$ \\
2   & $2.419$ & $\mathbf{1.000}$ & $2.247 \pm 0.014$ & $\mathbf{1.000}$     & $2.245 \pm 0.013$      & $+0.122$ \\
3   & $2.438$ & $\mathbf{1.000}$ & $2.250 \pm 0.015$ & $\mathbf{1.000}$     & $2.252 \pm 0.018$      & $-0.097$ \\
4   & $2.429$ & $\mathbf{1.000}$ & $2.238 \pm 0.013$ & $\mathbf{1.000}$     & $2.242 \pm 0.016$      & $-0.244$ \\
5   & $2.451$ & $\mathbf{1.000}$ & $2.255 \pm 0.025$ & $\mathbf{1.000}$     & $2.254 \pm 0.023$      & $+0.054$ \\
6   & $2.453$ & $\mathbf{1.000}$ & $2.247 \pm 0.016$ & $\mathbf{1.000}$     & $2.250 \pm 0.016$      & $-0.224$ \\
7   & $2.409$ & $\mathbf{1.000}$ & $2.247 \pm 0.017$ & $\mathbf{1.000}$     & $2.246 \pm 0.015$      & $+0.072$ \\
8   & $2.449$ & $\mathbf{1.000}$ & $2.244 \pm 0.011$ & $\mathbf{1.000}$     & $2.246 \pm 0.015$      & $-0.171$ \\
9   & $2.453$ & $\mathbf{1.000}$ & $2.258 \pm 0.014$ & $\mathbf{1.000}$     & $2.254 \pm 0.014$      & $+0.289$ \\
\bottomrule
\end{tabular}
\end{small}
\end{center}
\vskip -0.1in
\end{table}

\begin{table}[htb]
\caption{Information about each network trained on \textbf{CIFAR-10 with} dropout \textbf{before} pruning. All information as in table~\ref{tab:mnist-unpruned-no-dropout}. $p$-values less than $1/320$ and equal to $1$ are bolded.}
\label{tab:cifar-unpruned-dropout}
\vskip 0.15in
\begin{center}
\begin{small}
\begin{tabular}{lcccccccc}
\toprule
Run & N-cut   & $p$-value        & Dist.\ n-cuts     & $p$-value (non-zero) & Non-zero dist.\ n-cuts & Cohen's $d$ \\
\midrule
0   & $2.149$ & $\mathbf{0.003}$ & $2.233 \pm 0.002$ & $\mathbf{0.003}$     & $2.233 \pm 0.002$      & $+0.007$ \\
1   & $2.132$ & $\mathbf{0.003}$ & $2.234 \pm 0.002$ & $\mathbf{0.003}$     & $2.234 \pm 0.001$      & $-0.033$ \\
2   & $2.137$ & $\mathbf{0.003}$ & $2.234 \pm 0.003$ & $\mathbf{0.003}$     & $2.235 \pm 0.002$      & $-0.463$ \\
3   & $2.198$ & $\mathbf{0.003}$ & $2.238 \pm 0.002$ & $\mathbf{0.003}$     & $2.239 \pm 0.002$      & $-0.242$ \\
4   & $2.138$ & $\mathbf{0.003}$ & $2.230 \pm 0.002$ & $\mathbf{0.003}$     & $2.230 \pm 0.004$      & $+0.079$ \\
5   & $2.145$ & $\mathbf{0.003}$ & $2.234 \pm 0.002$ & $\mathbf{0.003}$     & $2.235 \pm 0.002$      & $-0.265$ \\
6   & $2.158$ & $\mathbf{0.003}$ & $2.234 \pm 0.005$ & $\mathbf{0.003}$     & $2.236 \pm 0.002$      & $-0.407$ \\
7   & $2.137$ & $\mathbf{0.003}$ & $2.233 \pm 0.002$ & $\mathbf{0.003}$     & $2.233 \pm 0.002$      & $+0.172$ \\
8   & $2.148$ & $\mathbf{0.003}$ & $2.235 \pm 0.002$ & $\mathbf{0.003}$     & $2.235 \pm 0.002$      & $-0.002$ \\
9   & $2.151$ & $\mathbf{0.003}$ & $2.235 \pm 0.003$ & $\mathbf{0.003}$     & $2.235 \pm 0.003$      & $+0.128$ \\
\bottomrule
\end{tabular}
\end{small}
\end{center}
\vskip -0.1in
\end{table}

\begin{table}[htb]
\caption{Information about each network trained on \textbf{CIFAR-10 without} dropout \textbf{after} pruning. All information as in table~\ref{tab:mnist-unpruned-no-dropout}. Note that each row corresponds to the network in the same row in table~\ref{tab:cifar-unpruned-no-dropout} after undergoing pruning. $p$-values less than $1/320$ and equal to $1$ are bolded.}
\label{tab:cifar-pruned-no-dropout}
\vskip 0.15in
\begin{center}
\begin{small}
\begin{tabular}{lcccccccc}
\toprule
Run & N-cut   & $p$-value        & Dist.\ n-cuts     & $p$-value (non-zero) & Non-zero dist.\ n-cuts & Cohen's $d$ \\
\midrule
0   & $2.049$ & $\mathbf{1.000}$ & $2.006 \pm 0.016$ & $0.751$              & $2.043 \pm 0.055$      & $-0.901$ \\
1   & $2.296$ & $\mathbf{1.000}$ & $1.999 \pm 0.016$ & $0.688$              & $2.354 \pm 0.327$      & $-1.533$ \\
2   & $2.037$ & $\mathbf{1.000}$ & $2.009 \pm 0.011$ & $0.907$              & $2.019 \pm 0.012$      & $-0.828$ \\
3   & $1.993$ & $0.377$          & $1.998 \pm 0.015$ & $0.688$              & $1.986 \pm 0.011$      & $+0.945$ \\
4   & $2.014$ & $0.844$          & $1.997 \pm 0.014$ & $0.907$              & $1.997 \pm 0.027$      & $+0.039$ \\
5   & $2.015$ & $0.657$          & $2.010 \pm 0.016$ & $0.252$              & $2.022 \pm 0.010$      & $-0.893$ \\
6   & $2.301$ & $\mathbf{1.000}$ & $1.995 \pm 0.017$ & $\mathbf{1.000}$     & $1.844 \pm 0.238$      & $+0.891$ \\
7   & $1.833$ & $\mathbf{0.003}$ & $1.997 \pm 0.013$ & $0.502$              & $1.898 \pm 0.404$      & $+0.347$ \\
8   & $2.054$ & $\mathbf{1.000}$ & $2.012 \pm 0.017$ & $\mathbf{1.000}$     & $2.021 \pm 0.013$      & $-0.585$ \\
9   & $2.008$ & $0.439$          & $2.008 \pm 0.014$ & $0.439$              & $2.012 \pm 0.008$      & $-0.371$ \\
\bottomrule
\end{tabular}
\end{small}
\end{center}
\vskip -0.1in
\end{table}

\begin{table}[htb]
\caption{Information about each network trained on \textbf{CIFAR-10 with} dropout \textbf{after} pruning. All information as in table~\ref{tab:mnist-unpruned-no-dropout}. Note that each row corresponds to the network in the same row in table~\ref{tab:cifar-unpruned-dropout} after undergoing pruning. $p$-values less than $1/320$ and equal to $1$ are bolded.}
\label{tab:cifar-pruned-dropout}
\vskip 0.15in
\begin{center}
\begin{small}
\begin{tabular}{lcccccccc}
\toprule
Run & N-cut   & $p$-value        & Dist.\ n-cuts     & $p$-value (non-zero) & Non-zero dist.\ n-cuts & Cohen's $d$ \\
\midrule
0   & $2.040$ & $\mathbf{1.000}$ & $1.990 \pm 0.009$ & $0.907$              & $1.873 \pm 0.128$      & $+1.298$  \\
1   & $1.827$ & $\mathbf{0.003}$ & $1.985 \pm 0.014$ & $0.252$              & $1.849 \pm 0.038$      & $+4.791$  \\
2   & $1.802$ & $\mathbf{0.003}$ & $1.990 \pm 0.012$ & $0.190$              & $1.833 \pm 0.050$      & $+4.311$  \\
3   & $1.915$ & $\mathbf{0.003}$ & $2.021 \pm 0.008$ & $\mathbf{0.003}$     & $1.939 \pm 0.033$      & $+3.420$  \\
4   & $1.733$ & $\mathbf{0.003}$ & $1.991 \pm 0.014$ & $\mathbf{0.003}$     & $1.767 \pm 0.015$      & $+15.043$ \\
5   & $1.780$ & $\mathbf{0.003}$ & $1.996 \pm 0.009$ & $0.065$              & $1.862 \pm 0.062$      & $+3.054$  \\
6   & $1.828$ & $\mathbf{0.003}$ & $2.003 \pm 0.007$ & $0.128$              & $1.936 \pm 0.128$      & $+0.741$  \\
7   & $1.853$ & $\mathbf{0.003}$ & $1.989 \pm 0.011$ & $0.844$              & $1.841 \pm 0.090$      & $+2.313$  \\
8   & $1.874$ & $\mathbf{0.003}$ & $1.998 \pm 0.009$ & $0.097$              & $1.985 \pm 0.100$      & $+0.174$  \\
9   & $1.750$ & $\mathbf{0.003}$ & $1.993 \pm 0.013$ & $0.097$              & $1.823 \pm 0.071$      & $+3.322$  \\
\bottomrule
\end{tabular}
\end{small}
\end{center}
\vskip -0.1in
\end{table}

%% file: appendix_sections/mixture_results.tex
The authors initially hypothesized that modularity is a result of different regions of the network processing different types of input. 
To test this, 
we developed mixture datasets composed of two original datasets.
These datasets are either of the `separate' type, where one original dataset has only classes 0 through 4 included and the other has classes 5 through 9 included; or the `overlapping' type, where both original datasets contribute examples of all classes. The datasets that we mix are MNIST, CIFAR-10, and LINES, which consists of 28$\times$28 images of white vertical lines on a black background, labeled with the number of vertical lines.

If modularity were a result of different regions of the network specializing in processing different types of information, we would expect that networks trained on mixture datasets would have n-cuts lower than those trained on either constituent dataset.
In fact, our results are ambiguous: for some datasets, the n-cuts of networks trained on the mixture datasets are lower than those trained on the constituent datasets, but for others, the n-cuts of networks trained on the mixture datasets are in-between those trained on the constituent datasets. As such, no particular conclusion can be drawn.

Table~\ref{tab:mixture-no-dropout} shows n-cuts and accuracies for networks trained without dropout, while table~\ref{tab:mixture-dropout} shows the same for networks trained with dropout. The broad pattern is that networks trained on mixtures between MNIST and CIFAR-10 have lower n-cuts than those trained on either individually, but networks trained on mixtures between LINES and another dataset have n-cuts intermediate between those trained on LINES and those trained on the other dataset. The one exception is that networks trained on LINES-CIFAR-10-SEP without dropout have lower n-cuts than those trained on either LINES without dropout and also those trained on CIFAR-10 without dropout. Since LINES is a very artificial dataset, it is possible that the results obtained for mixtures between MNIST and CIFAR-10 are more representative of other natural datasets.

\begin{table}[htb]
\caption{N-cuts and accuracies for networks trained without dropout. LINES-MNIST refers to the dataset where each class has data from both LINES and MNIST, while LINES-MNIST-SEP refers to the dataset where classes 0-4 have examples from LINES and classes 5-9 have examples from MNIST. Each row presents statistics for 10 networks. The ``N-cuts'' column shows the mean and standard deviation over 10 networks.}
\label{tab:mixture-no-dropout}
\vskip 0.15in
\begin{center}
\begin{small}
\begin{tabular}{lcccc}
\toprule
Dataset            & N-cuts            & Mean train acc.\ & Mean test acc.\ \\
\midrule
MNIST              & $2.000 \pm 0.035$ & $1.00 $          & $0.984$         \\
CIFAR-10           & $2.06  \pm 0.14 $ & $0.650$          & $0.415$         \\
LINES              & $1.599 \pm 0.049$ & $1.00 $          & $1.00 $         \\
LINES-MNIST        & $1.760 \pm 0.039$ & $1.00 $          & $0.991$         \\
LINES-CIFAR-10     & $1.632 \pm 0.032$ & $0.784$          & $0.708$         \\
MNIST-CIFAR-10     & $1.822 \pm 0.025$ & $0.805$          & $0.697$         \\
LINES-MNIST-SEP    & $1.842 \pm 0.046$ & $1.00 $          & $0.994$         \\
LINES-CIFAR-10-SEP & $1.536 \pm 0.033$ & $0.941$          & $0.817$         \\
MNIST-CIFAR-10-SEP & $1.587 \pm 0.041$ & $0.964$          & $0.810$         \\
\bottomrule
\end{tabular}
\end{small}
\end{center}
\vskip -0.1in
\end{table}

\begin{table}[htb]
\caption{N-cuts and accuracies for networks trained with dropout. Notation as in table~\ref{tab:mixture-no-dropout}.}
\label{tab:mixture-dropout}
\vskip 0.15in
\begin{center}
\begin{small}
\begin{tabular}{lcccc}
\toprule
Dataset            & N-cuts            & Mean train acc.\ & Mean test acc.\ \\
\midrule
MNIST              & $1.840 \pm 0.015$ & $0.967$          & $0.979$         \\
CIFAR-10           & $1.840 \pm 0.089$ & $0.427$          & $0.422$         \\
LINES              & $1.144 \pm 0.036$ & $0.913$          & $0.312$         \\
LINES-MNIST        & $1.272 \pm 0.056$ & $0.875$          & $0.635$         \\
LINES-CIFAR-10     & $1.418 \pm 0.031$ & $0.682$          & $0.459$         \\
MNIST-CIFAR-10     & $1.784 \pm 0.032$ & $0.714$          & $0.694$         \\
LINES-MNIST-SEP    & $1.517 \pm 0.020$ & $0.984$          & $0.885$         \\
LINES-CIFAR-10-SEP & $1.375 \pm 0.038$ & $0.828$          & $0.817$         \\
MNIST-CIFAR-10-SEP & $1.517 \pm 0.029$ & $0.853$          & $0.826$         \\
\bottomrule
\end{tabular}
\end{small}
\end{center}
\vskip -0.1in
\end{table}

%% file: appendix_sections/lesion_data.tex
\begin{figure}[thb]
\vskip 0.2in
\begin{center}
\centerline{\includegraphics[width=\textwidth]{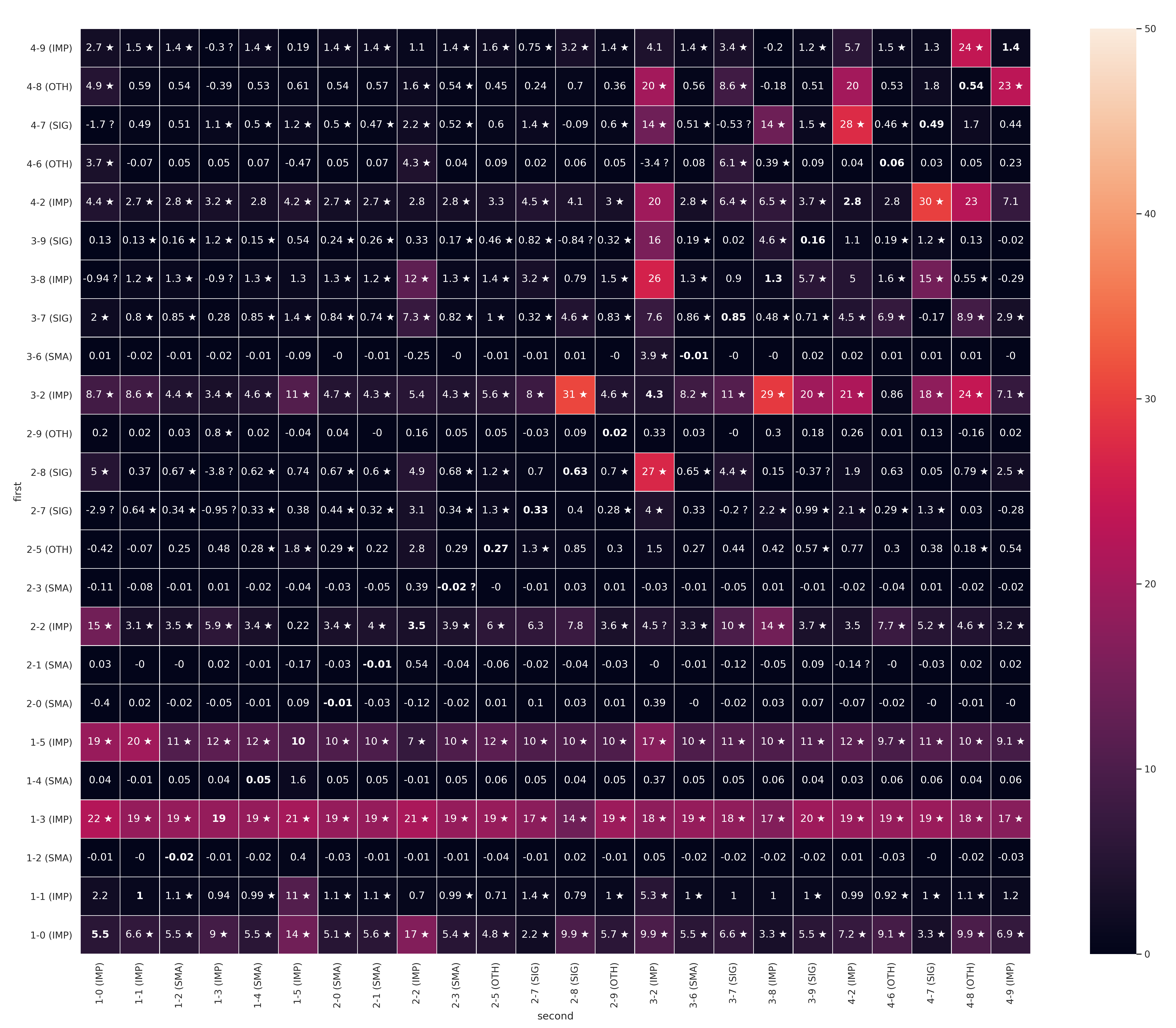}}
\caption{Dependency information for all pairs of sub-modules. Best viewed in color and zoomed in on a screen. Sub-modules are numbered by their layer and module number. They are also labeled by their importance: IMP stands for important, SMA stands for small, SIG stands for sig-but-not-diff, and OTH stands for other, terms which are defined in the caption of table~\ref{tab:single_lesion}. Cells on the diagonal are labeled by the accuracy damage in percentage points caused by zeroing out the corresponding sub-module in the single lesion experiments, and have bolded text, while cells off the diagonal are labeled by $100 \times \delta(\textrm{first},\,\textrm{second})$. Cells contain stars if the damage caused by lesioning the first is statistically significant at $p<0.02$ conditional on the second being lesioned, and question marks if the damage caused by lesioning the first was less than the damage caused by lesioning each of 50 random sets of neurons (i.e.\ $p=1.00$), conditional on the second being lesioned. Note that this plot includes sub-modules in the same layer, where the meaning is questionable since there can be no real dependency.}
\label{fig:double_lesion_table}
\end{center}
\vskip -0.2in
\end{figure}

Table~\ref{tab:single_lesion} shows data on the importance of sub-modules in the single lesion experiments, and is plotted in figure~\ref{fig:single_lesion_plot}. Table~\ref{tab:double-lesion-stats} shows the conditional importance data of important pairs of sub-modules that is plotted in figure~\ref{fig:dependency_info}~(a). The conditional importance data of all pairs of sub-modules is shown in figure~\ref{fig:double_lesion_table}.

\begin{table}[htb]
\caption{Table of sub-modules of a network trained on Fashion-MNIST using pruning and dropout. ``Acc.\ diff.'' means the difference in accuracy between the actual network and the network with that sub-module lesioned, while ``Acc.\ diff.\ dist.'' shows the mean and standard deviation of the distribution of accuracy differentials between the actual network and one with a random set of neurons lesioned. $p$-values are calculated using the method described by \citet{north2002note}. The ``Proportion'' column denotes the proportion of the layer's neurons that the sub-module represents. ``Important'' means that the damage caused by lesioning the sub-module was at least 1 percentage point and significant at $p<0.01$, ``sig-but-not-diff'' means that the lesioning damage was significant but less than 1 percentage point, ``other'' meant that the lesioning damage was not significant at $p<0.01$ and also less than 1 percentage point, while ``small'' means that the neurons of the sub-module constitute less than 5\% of the layer.}
\label{tab:single_lesion}
\vskip 0.15in
\begin{center}
\begin{small}
\begin{tabular}{llcccccc}
\toprule
Layer & Label & Acc.\ diff.\ & $p$-value & Proportion & Type             & Acc.\ diff.\ dist.\     \\
\midrule                                                                                           
1     & 0     & $-0.0552$    & $0.0099$  & $0.250$    & important        & $-0.0034  \pm 0.0026 $  \\
1     & 1     & $-0.0105$    & $0.0099$  & $0.123$    & important        & $-0.0011  \pm 0.0015 $  \\
1     & 2     & $+0.0002$    & $0.64  $  & $0.015$    & small            & $-0.00007 \pm 0.00057$  \\
1     & 3     & $-0.1870$    & $0.0099$  & $0.245$    & important        & $-0.0037  \pm 0.0031 $  \\
1     & 4     & $-0.0005$    & $0.38  $  & $0.049$    & small            & $-0.0002  \pm 0.0010 $  \\
1     & 5     & $-0.1024$    & $0.0099$  & $0.319$    & important        & $-0.0052  \pm 0.0030 $  \\
\midrule
2     & 0     & $+0.0001$    & $0.52  $  & $0.020$    & small            & $+0.00001  \pm 0.00051$ \\
2     & 1     & $+0.0001$    & $0.55  $  & $0.012$    & small            & $+0.00001  \pm 0.00052$ \\
2     & 2     & $-0.0350$    & $0.0099$  & $0.461$    & important        & $-0.0042   \pm 0.0034 $ \\
2     & 3     & $+0.0002$    & $0.71  $  & $0.008$    & small            & $-0.000003 \pm 0.00031$ \\
2     & 5     & $-0.0027$    & $0.020 $  & $0.109$    & other            & $-0.0004   \pm 0.0010 $ \\
2     & 7     & $-0.0033$    & $0.0099$  & $0.098$    & sig-but-not-diff & $-0.00030  \pm 0.00095$ \\
2     & 8     & $-0.0063$    & $0.0099$  & $0.242$    & sig-but-not-diff & $-0.0013   \pm 0.0016 $ \\
2     & 9     & $-0.0002$    & $0.39  $  & $0.051$    & other            & $-0.00006  \pm 0.00074$ \\
\midrule
3     & 2     & $-0.0433$    & $0.0099$  & $0.527$    & important        & $+0.0019  \pm 0.0021 $  \\
3     & 6     & $+0.0001$    & $0.55  $  & $0.004$    & small            & $-0.00007 \pm 0.00027$  \\
3     & 7     & $-0.0085$    & $0.0099$  & $0.113$    & sig-but-not-diff & $+0.0011  \pm 0.0010 $  \\
3     & 8     & $-0.0127$    & $0.0099$  & $0.234$    & important        & $+0.0015  \pm 0.0011 $  \\
3     & 9     & $-0.0016$    & $0.0099$  & $0.121$    & sig-but-not-diff & $+0.00096 \pm 0.00089$  \\
\midrule
4     & 2     & $-0.0279$    & $0.0099$  & $0.414$    & important        & $-0.0076  \pm 0.0035 $  \\
4     & 6     & $-0.0006$    & $0.69  $  & $0.082$    & other            & $-0.00104 \pm 0.00072$  \\
4     & 7     & $-0.0049$    & $0.0099$  & $0.125$    & sig-but-not-diff & $-0.0016  \pm 0.00083$  \\
4     & 8     & $-0.0054$    & $0.059 $  & $0.219$    & other            & $-0.0029  \pm 0.0014 $  \\
4     & 9     & $-0.0136$    & $0.0099$  & $0.160$    & important        & $-0.0019  \pm 0.0010 $  \\
\bottomrule
\end{tabular}
\end{small}
\end{center}
\vskip -0.1in
\end{table}


\begin{table}[tbh]
\caption{Dependency information of pairs of important sub-modules of an MLP trained on Fashion-MNIST with dropout. $X$ and $Y$ represent sub-modules, $X$ being the one earlier in the network. They are numbered by first their layer and then their module number. $X|Y$ means whether $X$ is important conditioned on $Y$, and vice versa. $\delta(X,Y)$ as well as ``importance'' are defined in subsection~\ref{subsec:double_lesion}.}
\label{tab:double-lesion-stats}
\vskip 0.15in
\begin{center}
\begin{small}
\begin{tabular}{cccccc}
\toprule
$X$ & $Y$ & $X|Y$    & $Y|X$    & $\delta(X,Y)$ & $\delta(Y,X)$ \\
\midrule
1-0 & 2-2 & $\surd$  & $\surd$  & $+0.166$      & $+0.146$      \\
1-0 & 3-2 & $\surd$  & $\surd$  & $+0.099$      & $+0.087$      \\
1-0 & 3-8 & $\surd$  & $\times$ & $+0.033$      & $-0.009$      \\
1-0 & 4-2 & $\surd$  & $\surd$  & $+0.072$      & $+0.044$      \\
1-0 & 4-9 & $\surd$  & $\surd$  & $+0.069$      & $+0.027$      \\
1-1 & 2-2 & $\times$ & $\surd$  & $+0.007$      & $+0.031$      \\
1-1 & 3-2 & $\surd$  & $\surd$  & $+0.053$      & $+0.086$      \\
1-1 & 3-8 & $\times$ & $\surd$  & $+0.010$      & $+0.012$      \\
1-1 & 4-2 & $\times$ & $\surd$  & $+0.010$      & $+0.027$      \\
1-1 & 4-9 & $\times$ & $\surd$  & $+0.012$      & $+0.015$      \\
1-3 & 2-2 & $\surd$  & $\surd$  & $+0.211$      & $+0.059$      \\
1-3 & 3-2 & $\surd$  & $\surd$  & $+0.178$      & $+0.034$      \\
1-3 & 3-8 & $\surd$  & $\times$ & $+0.165$      & $-0.009$      \\
1-3 & 4-2 & $\surd$  & $\surd$  & $+0.191$      & $+0.032$      \\
1-3 & 4-9 & $\surd$  & $\times$ & $+0.170$      & $-0.003$      \\
1-5 & 2-2 & $\surd$  & $\times$ & $+0.070$      & $+0.002$      \\
1-5 & 3-2 & $\surd$  & $\surd$  & $+0.168$      & $+0.109$      \\
1-5 & 3-8 & $\surd$  & $\times$ & $+0.103$      & $+0.013$      \\
1-5 & 4-2 & $\surd$  & $\surd$  & $+0.117$      & $+0.042$      \\
1-5 & 4-9 & $\surd$  & $\times$ & $+0.091$      & $+0.002$      \\
2-2 & 3-2 & $\times$ & $\times$ & $+0.045$      & $+0.054$      \\
2-2 & 3-8 & $\surd$  & $\surd$  & $+0.145$      & $+0.123$      \\
2-2 & 4-2 & $\times$ & $\times$ & $+0.035$      & $+0.028$      \\
2-2 & 4-9 & $\surd$  & $\times$ & $+0.032$      & $+0.011$      \\
3-2 & 4-2 & $\surd$  & $\times$ & $+0.214$      & $+0.199$      \\
3-2 & 4-9 & $\surd$  & $\times$ & $+0.071$      & $+0.041$      \\
3-8 & 4-2 & $\times$ & $\surd$  & $+0.050$      & $+0.065$      \\
3-8 & 4-9 & $\times$ & $\times$ & $-0.003$      & $-0.002$      \\
\bottomrule
\end{tabular}
\end{small}
\end{center}
\vskip -0.1in
\end{table}

